\documentclass[sn-basic,iicol,sns-apa]{sn-jnl}

\usepackage{array}
\usepackage{graphicx}%
\usepackage{multirow}%
\usepackage{amsmath,amssymb,amsfonts}%
\usepackage{amsthm}%
\usepackage{mathrsfs}%
\usepackage[title]{appendix}%
\usepackage{xcolor}%
\usepackage{textcomp}%
\usepackage{manyfoot}%
\usepackage{booktabs}%
\usepackage{algorithm}%
\usepackage{algorithmicx}%
\usepackage{algpseudocode}%
\usepackage{listings}%
\usepackage{todonotes}
\usepackage{subfigure}
\usepackage{arydshln}
\usepackage{pifont}

\usepackage{comment}

\newcommand{\cmark}{\ding{51}}%
\newcommand{\xmark}{\ding{55}}%

\begin{document}
\title[Hyperspectral Benchmark: Bridging the Gap between HSI Applications through Comprehensive Dataset and Pretraining]{Hyperspectral Benchmark: Bridging the Gap between HSI Applications through Comprehensive Dataset and Pretraining}


\author*[]{\fnm{Hannah} \sur{Frank}}\email{hannah.frank@uni-tuebingen.de}
\equalcont{These authors contributed equally to this work.}

\author[]{\fnm{Leon Amadeus} \sur{Varga}}\email{leon.varga@uni-tuebingen.de}
\equalcont{These authors contributed equally to this work.}

\author[]{\fnm{Andreas} \sur{Zell}}\email{andreas.zell@uni-tuebingen.de}

\affil[]{\orgdiv{Cognitive Systems}, \orgname{University of Tuebingen}, \orgaddress{\street{Sand 1}, \city{Tuebingen}, \postcode{72076}, \state{Baden-Württemberg}, \country{Germany}}}

\abstract{Hyperspectral Imaging (HSI) serves as a non-destructive spatial spectroscopy technique with a multitude of potential applications. However, a recurring challenge lies in the limited size of the target datasets, impeding exhaustive architecture search. Consequently, when venturing into novel applications, reliance on established methodologies becomes commonplace, in the hope that they exhibit favorable generalization characteristics. Regrettably, this optimism is often unfounded due to the fine-tuned nature of models tailored to specific HSI contexts.

To address this predicament, this study introduces an innovative benchmark dataset encompassing three markedly distinct HSI applications: food inspection, remote sensing, and recycling. This comprehensive dataset affords a finer assessment of hyperspectral model capabilities. Moreover, this benchmark facilitates an incisive examination of prevailing state-of-the-art techniques, consequently fostering the evolution of superior methodologies.

Furthermore, the enhanced diversity inherent in the benchmark dataset underpins the establishment of a pretraining pipeline for HSI. This pretraining regimen serves to enhance the stability of training processes for larger models. Additionally, a procedural framework is delineated, offering insights into the handling of applications afflicted by limited target dataset sizes.}




\maketitle
\section{Introduction and Related Work}
\label{sec:intro}

Hyperspectral imaging (HSI) is a non-destructive measurement technique, which can be seen as a spatial spectroscopy. The sensors cover a wide spectrum beyond the visible light, and usually provide hundreds of narrow bands. The recording of a hyperspectral camera is called a hyperspectral cube, with two spatial dimensions (x and y) and a channel dimension ($\lambda$). Each pixel is a high-dimensional vector whose channel entries correspond to the intensity in a specific wavelength.

With the advantage of capturing additional spectral information, the hyperspectral imaging technology became increasingly popular and has been applied in many fields. In remote sensing where HSI has its origin (e.g., \cite{ChenLZWG14}), but also in medical applications by \cite{LuF14}, in agriculture (``precision farming") by \cite{LuDLHS20},  in the recycling industry (e.g., \cite{BonifaziCSP19}), and in the food industry, where the HSI technique is used, for example, to measure the quality or ripeness of fruit (e.g., \cite{GirodLDOSG08, VargaMZ21}). The recordings and the important features of these applications differ widely.

The interpretation of a hyperspectral cube poses challenges for human experts, which lead to the development of various processing methods. However, these solutions are often tailored to specific applications, making adaptation to new scenarios difficult. Ideally, a method that demonstrates strong generalization capabilities and can be readily applied to diverse data sets and tasks without significant modifications would be desired. To validate existing approved approaches and novel models based on these criteria, a comprehensive benchmark testing different application scenarios is required.

In this work, we focus on the classification and segmentation task of hyperspectral recordings, as these are still the most common applications of HSI. More complex computer vision tasks, like object detection or object tracking, are with the most current hyperspectral cameras not practicable. For classification, the task is to determine the class of a sample in a recording. In contrast, segmentation produces a segmentation mask for a recording, which indicates the class of each pixel.

In the early stages, classical machine learning (ML) approaches, like
support vector machines (SVM) described by \cite{DBLP:books/daglib/0026018} 
were used for HSI classification.
Other methods focused on feature extraction or dimensionality reduction, e.g., via principal component analysis (PCA) proposed by \cite{doi:10.1080/14786440109462720} as a preprocessing step. These methods operate pixelwise and therefore on the spectral dimension only. However, it has been shown that HSI classification performance is highly dependent on both, spatial and spectral information (e.g., \cite{Yan10}).

Nowadays, hyperspectral image data is mostly evaluated using deep learning.
Deep learning models, like 
autoencoders (AE) \citep{LeCun87}, 
recurrent neural networks (RNN) \citep{RumelhartHW86}, 
or convolutional neural networks (CNN) \citep{Fukushima80},
have been applied for HSI classification (e.g., \cite{LinCZW13, MouGZ17, YangYLLZH18}).

Conventional CNNs use 2D convolutions that are applied over the spatial dimensions, but are not able to handle the spectral information.
3D convolutions, on the other hand, can extract the spectral and spatial information simultaneously, but at the cost of increased computational complexity.
Recently, there have been attempts to combine both, 2D and 3D convolutions, in order to benefit from the spatial and spectral feature learning capability, by simultaneously overcoming this drawback. Such networks are then referred to as hybrid or 3D-2D CNNs (e.g., \cite{RoyKDC20}).

An alternative method for simultaneously leveraging spatial and spectral information is to apply 2D convolutions on transformed data. For instance, \cite{Chakraborty2021SpectralNETES} employed wavelet transformations to incorporate both aspects.

\begin{figure*}
    \centering
    \subfigure[]{\includegraphics[width=.3\textwidth]{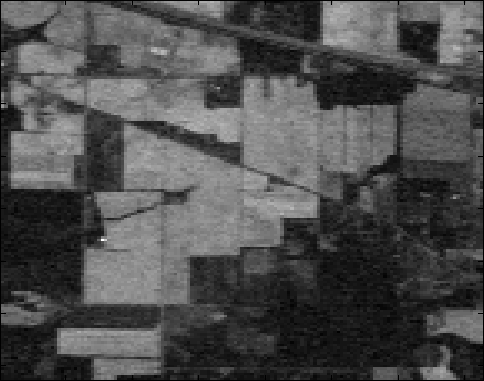}\label{fig:indian_pines_avg}}
   \hspace{50pt} 
    \subfigure[]{\includegraphics[width=.3\textwidth]{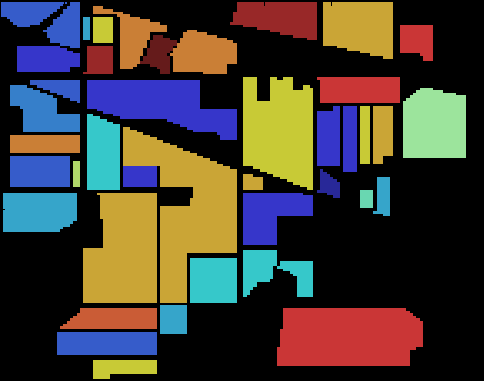}\label{fig:indian_pines_gt}}
    \caption{Indian Pines hyperspectral recording from the Hyperspectral Remote Sensing Scenes (HRSS) data set. (a) depicts the average of the 200 bands. (b) illustrates the corresponding ground truth segmentation mask.}
    \label{fig:indian_pines}
\end{figure*}
 
Although CNNs have proven to be powerful in extracting spatial and locally contextual information, they fail to capture the globally sequential information, especially long-term dependencies in the spectral dimension of the hyperspectral data. vision transformers (ViT), as described by \cite{DosovitskiyBKWZUDMHGUH20} are specifically designed for and very efficient at analyzing sequential data, mainly because of the self-attention mechanism.
Thus, ViTs have also been successfully applied for HSI classification (e.g., \cite{HeZYZL20, HongHYGZPC22, YangCLZ22}).

A couple of publications reviewed the current development of HSI models \citep{PaolettiHPP19, LiSFCGB19, AhmadSRHWYKMDC22}. However, these reviews often lack the most recent developments, including vision transformers, and focus on remote sensing scenes only.

For the assessment of hyperspectral models, the established hyperspectral remote sensing scenes (HRSS) dataset published by \cite{HRSSdatasets} is the primary choice, despite its limitations (e.g., lack of a defined training-test split), which will be discussed in Section \ref{sec:data}. This dataset comprises satellite recordings with the objective of ground classification as a segmentation task. Although there are different recorded scenarios, the application is very specific and differs widely from the much more common inline inspection. Consequently, recent model developments' conclusions are not directly transferable to other hyperspectral applications.

To address this, we introduce a comprehensive framework to evaluate and compare various approaches and models for different hyperspectral imaging (HSI) classification tasks. The benchmark, see Section \ref{sec:data}, consists of three different data sets, provides model implementations for 23 models, and allows a simple integration of additional ones. Through the establishment of consistent training and evaluation protocols, we facilitate equitable model comparisons and enhance result reproducibility in the field of HSI.

Furthermore, leveraging the proposed framework, we conduct an in-depth analysis of current state-of-the-art models for HSI, yielding valuable insights, as presented in Section \ref{sec:results}.

Finally, we explore the technique of pretraining, where models are initially trained on large datasets with disparate applications, and subsequently fine-tuned on the target dataset, which is typically smaller. This pretraining process stabilizes training and leads to notably improved results for color images (see, e.g., \cite{KrizhevskySH12}). Similar results were produced for the HRSS data set with different scenes \citep{LeeEK22, WindrimMMCR18}. Still, this is very specific for the remote sensing use case. In Section \ref{sec:pretraining}, we leverage the proposed benchmark datasets to evaluate pretraining's efficacy for various hyperspectral applications and highlight essential considerations for its successful implementation.


\section{Proposed Benchmark}\label{sec:data}
As discussed in the introduction, the current data sets serve well for developing models tailored to specific applications. However, this approach becomes limiting when attempting to address entirely new applications, as the fine-tuned models struggle to generalize effectively to unseen scenarios.

The primary objective of the proposed benchmark is to overcome this limitation by amalgamating data sets from multiple applications. By doing so, we aim to evaluate the performance of models across diverse hyperspectral imaging (HSI) applications, providing a comprehensive measure of their overall ability to handle hyperspectral recordings.

This section presents a detailed description of the published benchmark. It comprises three carefully chosen data sets, selected based on their widespread usage within the community, substantial size, and diverse application requirements.

The entire benchmark data set has a size of approximately 250 Gigabytes, and we have made it easily accessible for download through the provided download script at \url{https://github.com/cogsys-tuebingen/hsi_benchmark}. Further, a PyTorch framework is provided, which allows reproducing our results and straightforward implementation of additional models.

\begin{figure*}
    \centering
    \def\arraystretch{1.5}
    \subfigure[]{\includegraphics[width=.2\textwidth]{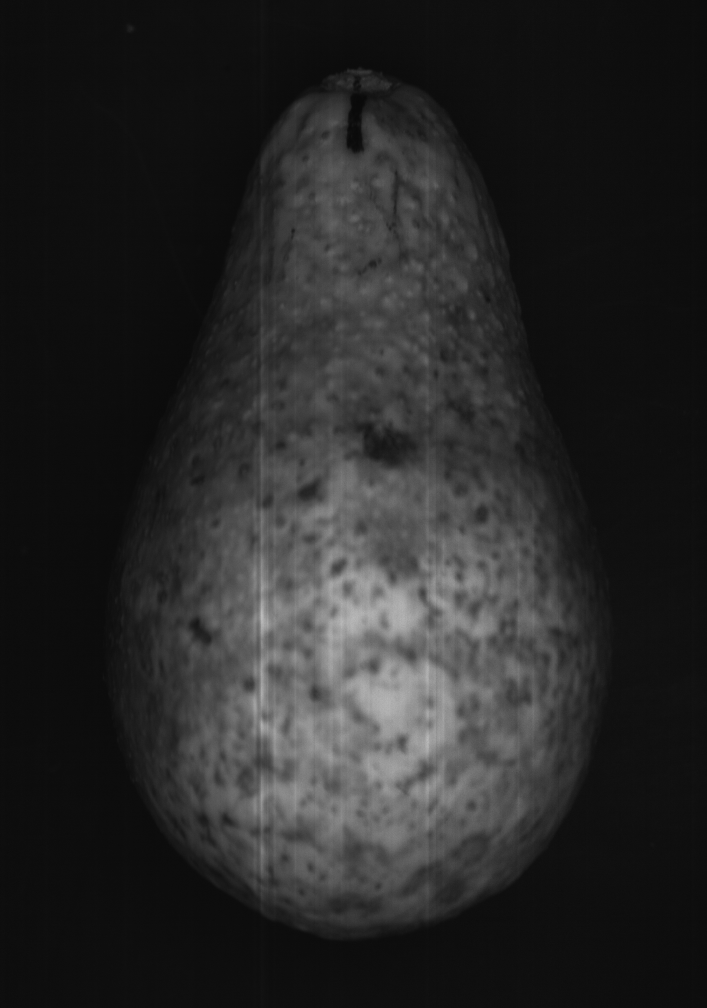}\label{fig:avocado_avg}}
   \hspace{50pt} 
    \subfigure[]{\includegraphics[width=0.18\textwidth]{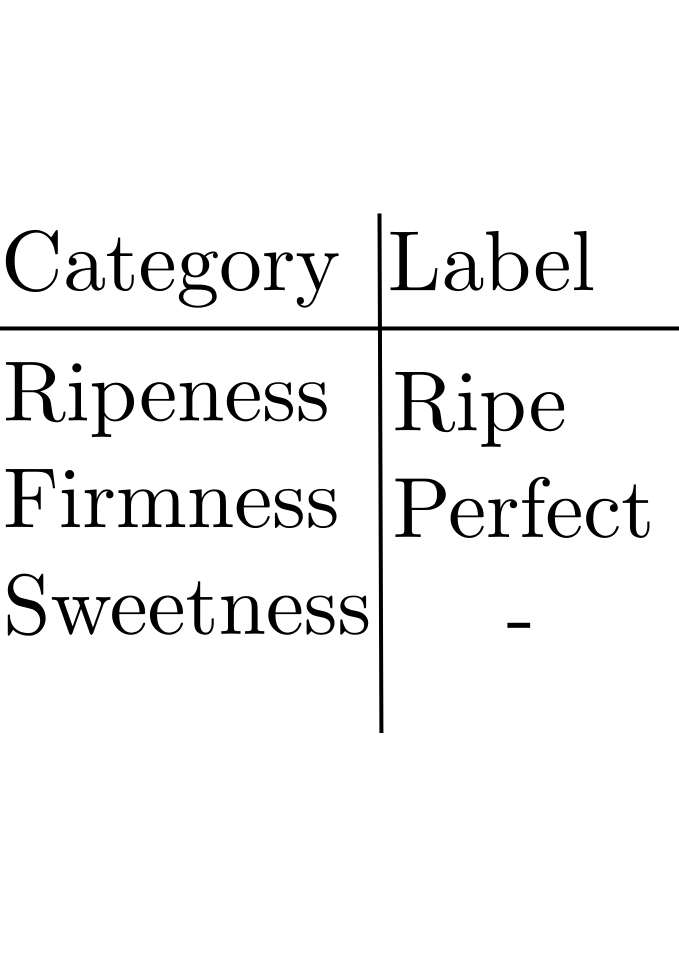}
    \label{fig:avocado_gt}}
    \caption{An avocado recording of the DeepHS Fruit data set. (a) depicts the average of the bands. (b) shows the corresponding ground truth labels.}
    \label{fig:avocado}
\end{figure*} 

\subsection{Hyperspectral Remote Sensing Scenes}
M. Graña et al. collected satellite recordings and provide these as the hyperspectral remote sensing scenes (HRSS) data set \citep{HRSSdatasets}. This data set is currently the most established data set for the evaluation of hyperspectral models \citep{PaolettiHPP19, LiSFCGB19, AhmadSRHWYKMDC22, RoyKDC20, Chakraborty2021SpectralNETES, HeZYZL20, HongHYGZPC22, YangCLZ22}.

Within the HRSS data set, several hyperspectral recordings are available for conducting segmentation tasks, with each recording being evaluated independently. Each of these recordings comes accompanied by a ground truth segmentation mask, facilitating precise evaluation. However, it's worth noting that the different recordings within the data set do not share the same classes. So, it is not straightforward to combine all recordings in a single evaluation. 

For the benchmark, we select the three most frequently used recordings. 
\begin{itemize}
    \item \textit{Indian Pines}:
    The Indian Pines scene (shown in Fig. \ref{fig:indian_pines}) contains two-thirds agriculture, and one-third forest or other natural perennial vegetation. The task is to distinguish crop types. The corrected version without the noisy water absorption bands is used. The recording contains 16 classes.
    ($x=145; y=145; \lambda=220$, AVIRIS sensor)
    \item \textit{Salinas}:
    The Salinas scene shows the Salinas Valley in California. Again, different crop types should be detected. The corrected version of the Salinas data set is used. It contains 16 classes.
    ($x=512; y=217; \lambda=224$, AVIRIS sensor)   
    \item \textit{Pavia University}:
    The recording Pavia University shows the university of Pavia, Italy. It covers nine classes, which represent the	nature of the ground (e.g. asphalt, bitumen, soil).
    ($x=610; y=340; \lambda=102$, ROSIS sensor)   
\end{itemize}
For detailed specifications of the sensors, refer to Table \ref{tab:cameras}.

Despite the widespread usage of the HRSS data set, it lacks an established training-validation-test split. Oftentimes, only the training-test ratio is reported, leading to complexities in reproducing experiments and making fair model comparisons. The absence of a standardized partitioning scheme hinders the ability to assess model performance consistently and hampers the reproducibility of results. As a consequence, researchers encounter challenges in accurately replicating experiments and drawing meaningful comparisons between different models. Establishing a well-defined training-validation-test split is crucial for enhancing the rigor and reliability of research outcomes and fostering advancements in this area.

To solve this issue, we define fixed training-validation-test splits independent of the train-test ratios and with balanced classes (see Appendix \ref{app:hrss_split}). The train-test-ratio defines the fraction of the image used for training. We define two train-test-ratios (10 \% and 30 \% of the pixels for training, and the remaining pixels for testing), as these are the most commonly used. Although evaluations involving smaller train-test ratios are conducted, they are considered supplementary and are not part of the main evaluation. By employing fixed splits and accounting for class balance, we ensure consistent reproducibility and fostering robust comparisons between different models.

\begin{figure*}
    \centering
    \def\arraystretch{1.5}
    \subfigure[]{\includegraphics[width=.3\textwidth]{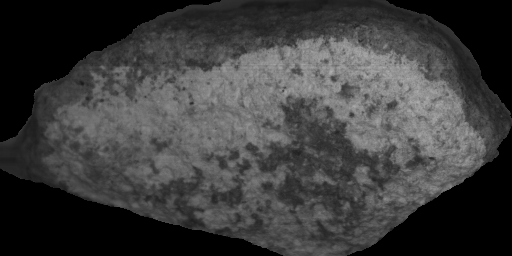}\label{fig:brick_avg}}
   \hspace{50pt} 
    \subfigure[]{\includegraphics[width=0.3\textwidth]{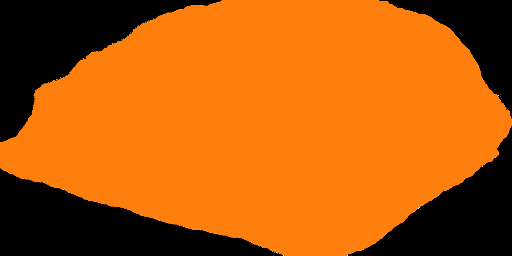}
    \label{fig:brick_gt}}
    \caption{A brick recording of the DeepHS Debris data set. (a) depicts the average of the bands. (b) shows the corresponding segmentation mask. Additionally, the ground-truth label is 'brick' as only a single object is shown in the recording.}
\end{figure*}

\subsection{DeepHS Fruit}
DeepHS Fruit version 2 \citep{VargaMZ21, VargaFZ23} presents a classification data set tailored for fruit ripeness prediction through hyperspectral imaging. The data set comprises approximately 5000 recordings of individual fruit specimens belonging to five distinct fruit types, namely avocado, kaki, kiwi, mango, and papaya. To ensure high accuracy and precision, the recordings were conducted under controlled laboratory conditions using three advanced hyperspectral cameras.

Specifically, two cameras, namely the Specim FX10 and Corning microHSI 410 Vis-NIR Hyperspectral Sensor, were utilized to record high-resolution images, capturing spectral information within the visible range and the lower near-infrared range. In addition, a near-infrared camera, the Innospec RedEye, was employed to acquire complementary data, further enhancing the data set's richness and reliability. For detailed specifications of the cameras, refer to Table \ref{tab:cameras}.

The inclusion of hyperspectral cubes, along with elaborately measured ground truth labels, offers valuable insights into fruit ripeness. The data set's labels encompass three key attributes per recording, namely:
\begin{itemize}
\item \textbf{Ripeness}, encompassing the distinct stages of unripe, ripe, and overripe fruit.
\item \textbf{Sweetness}, characterized by discerning between not sweet, sweet, and overly sweet fruit.
\item \textbf{Firmness}, judiciously classified as too firm, perfect, or too soft.
\end{itemize}

This data set provides a robust foundation for objectwise classification tasks with three distinct categories, thereby serving as a benchmark for research and application endeavors within the domain of fruit ripeness prediction using hyperspectral imaging. The original publication's definition of training-validation-test sets has been adopted for consistency and comparability.

\subsection{DeepHS Debris}
The DeepHS Debris data set is a contribution within this publication. Similar to the DeepHS Fruit data set, it was recorded under laboratory conditions. The data set encompasses both an objectwise classification task and a segmentation task, with the primary objective of distinguishing components of construction waste. Over a hundred samples of five common debris types (asphalt, brick, ceramic, concrete and tile) were recorded. 

The acquisition was performed with two hyperspectral cameras (Specim FX10 and Corning microHSI 410 Vis-NIR Hyperspectral Sensor). Both record the visible range and the lower near-infrared (up to 1000 nm) and produced high-resolution images of the debris under consideration. 
Again, refer to Table \ref{tab:cameras} for detailed specifications of the cameras.

In contrast to the previously mentioned data sets, the DeepHS Debris data set uniquely offers both an objectwise classification track and a segmentation track. This allows the evaluation of models for both use-cases. Both tracks are evaluated separately. To ensure consistency and comparability in research endeavors, a fixed training-validation-test split has been defined for this data set.

\begin{table*}[]
    \centering
    \def\arraystretch{1.5}
    \begin{tabular}{l|l|l|l}
        \hline
         Camera & Wavelength range& Application & Data set \\ \hline
         AVIRIS sensor & $400 - 2500 nm$ (224 bands) & Satellite & HRSS - Indian Pines and Salinas \\
         ROSIS sensor & $430 - 860 nm$ (115 bands) & Satellite & HRSS - University of Pavia \\
         Corning microHSI 410 & $408 - 901 nm$ (249 bands) & Inline & DeepHS Fruit and Debris\\
         Innospec RedEye & $920 - 1730 nm$ (252 bands) & Inline & DeepHS Fruit \\
         Specim FX10 & $400 - 1000 nm$ (224 bands) & Inline & DeepHS Fruit and Debris \\
        \hline
    \end{tabular}
    \caption{List of all used hyperspectral cameras and their specifications.}
    \label{tab:cameras}
\end{table*}

\subsection{DeepHS Benchmark}
The proposed hyperspectral benchmark was created by amalgamating the abovementioned three distinct data sets, each representing significant and emerging applications of hyperspectral imaging (HSI): remote sensing, food inspection, and recycling. The inclusion of these diverse applications in the benchmark ensures its relevance and importance in the field.

Each selected data set possesses unique characteristics and demands specific requirements from models. The HRSS data set focuses on segmentation, emphasizing the significance of texture features. In contrast, DeepHS Fruit revolves around objectwise classification, where spectral identification takes precedence. Lastly, DeepHS Debris allows both objectwise classification and pixelwise segmentation, encompassing the tasks of the other data sets. Despite their dissimilarities, all three data sets carry equal weight in our evaluation.

Through the integration of these data sets, the benchmark offers a comprehensive assessment of models' performance in hyperspectral applications. This evaluation enables us to make meaningful observations about the models' generalizability and provides valuable insights for future advancements in model development. 

In Appendix \ref{app:split}, a comprehensive outline of the dataset configurations employed for the benchmark is furnished. The table therein not only enumerates the specific configurations but also furnishes the dimensions of the training, validation, and test sets corresponding to each configuration.

By introducing this benchmark, we facilitate the pretraining of hyperspectral models, a practice widely employed in neural networks for color image applications (see, e.g. \cite{KrizhevskySH12}). Traditionally, large data sets are utilized for pretraining, allowing models to assimilate crucial features beforehand. Consequently, fine-tuning on a smaller target data set, specific to the actual application, becomes more straightforward.
While some publications have utilized the HRSS data set for pretraining and fine-tuning exclusively within the realm of remote sensing recordings \citep{LeeEK22, WindrimMMCR18}, this approach lacks versatility for other applications. This limitation and the usage of the proposed benchmark for pretraining is explored further in Section \ref{sec:pretraining}.

\section{Experiments}
In this section, the baseline experiments for the proposed data sets are described. After introducing the selected models with their specific training parameters, the experiment setups with the training procedures are presented. Finally, the results of the experiments are discussed and conclusions helping future model developments are provided.

\subsection{Models}
\label{sec:models}

\begin{table*}[]
    \centering
    \def\arraystretch{1.5}
    \begin{tabular}{l|p{2.5cm}crr}
        \hline
         Model name & Type &  \begin{tabular}{@{}c@{}}Spatial \\ context\end{tabular} & Input &  \begin{tabular}{@{}c@{}}Trainable \\ parameters\end{tabular} \\ \hline
         SVM \cite{DBLP:books/daglib/0026018} & \multirow{2}{*}{Classical ML} & \xmark & PCA(10)& -\\
         PLS-DA \cite{barker2003partial} & & \xmark & Raw & - \\ 
         \hdashline
         MLP \cite{PaolettiHPP19}& \multirow{2}{*}{\begin{tabular}{@{}l@{}}
                Basic\\
                neural networks
         \end{tabular}} & \xmark & Raw & 29,000\\
         RNN \cite{RumelhartHW86} & & \xmark & Raw & 27,000\\ 
         \hdashline
         1D CNN \cite{PaolettiHPP19} & \multirow{5}{*}{\begin{tabular}{@{}l@{}}Convolutional\\ neural\\ networks\end{tabular}} & \xmark & Raw & 73,000\\ 
         2D CNN \cite{PaolettiHPP19} &  & \cmark & PCA(40) & 7,500,000\\
         2D CNN (spatial)\cite{PaolettiHPP19} & & \cmark & Mean & 7,500,000\\
         2D CNN (spectral) & & \xmark & Raw& 7,500,000\\
         3D CNN \cite{PaolettiHPP19} & & \cmark & PCA(40) & 29,000,000\\ 
         \hdashline
         Gabor CNN \cite{GhamisiMLSTM18}& \multirow{2}{*}{CNNs + Filter} & \cmark & PCA(3) & 7,400,000\\
         EMP CNN \cite{GhamisiMLSTM18}& & \cmark & PCA(3) & 7,500,000\\ 
         \hdashline
         ResNet-18 \cite{HeZRS16}& \multirow{4}{*}{ResNet networks} & \cmark & Raw & 12,000,000\\
         ResNet-152 \cite{HeZRS16}& & \cmark & Raw & 59,000,000\\
         ResNet-18+HyveConv & & \cmark & Raw & 11,000,000\\
         ResNet-152+HyveConv & & \cmark & Raw & 58,000,000\\ 
         \hdashline
         DeepHS-Net \cite{VargaMZ21}& \multirow{3}{*}{DeepHS networks} & \cmark & Raw & 31,000\\
         DeepHS-Net+HyveConv \cite{VargaMBZ23}& & \cmark & Raw & 17,000\\ 
         DeepHS-Hybrid-Net \cite{VargaFZ23}& & \cmark & Raw & 1,300,000\\
         \hdashline
         SpectralNET \cite{Chakraborty2021SpectralNETES} & \multirow{2}{*}{\begin{tabular}{@{}l@{}}
                State-of-the-art\\
                for HRSS
         \end{tabular}}& \cmark & Raw & 8,300,000\\
         HybridSN \cite{RoyKDC20}& & \cmark & PCA(30) & 50,000,000\\ 
         \hdashline
         Attention-based CNN \cite{LorenzoTMN20} & \multirow{3}{*}{\begin{tabular}{@{}l@{}}
              Attention-based  \\
              approaches\end{tabular}} & \xmark & Raw & 2,000,000\\
         SpectralFormer \cite{HongHYGZPC22} & & \cmark & Raw & 1,000,000\\
         HiT \cite{YangCLZ22} & & \cmark & Raw & 59,000,000\\
        \hline
    \end{tabular}
    \caption{Model overview: These models are used for the analysis and provide baseline results for the proposed benchmark}
    \label{tab:model_overview}
\end{table*}

A selection of diverse models was chosen to encompass both state-of-the-art techniques for hyperspectral image classification and contemporary approaches for handling the hyperspectral cube. The complete list of models can be found in Table \ref{tab:model_overview} and will be elaborated upon in the subsequent paragraphs. Unless otherwise specified, the default configuration suggested in the original publications of each model was employed. 

As visualized in Tab. \ref{tab:model_overview}, the 23 selected models can be assigned to eight different groups, which are based on fundamentally different techniques to handle the hyperspectral data. 

The first category comprises classical machine learning techniques, namely support vector machine (SVM) \citep{DBLP:books/daglib/0026018} and partial least-squares discriminant analysis (PLS-DA) \citep{barker2003partial}. The SVM separates input data using hyperplanes and can handle non-linearly separable spaces by employing the kernel trick. In our case, we utilize the radial-basis function kernel for its robustness. PLS-DA, similar to principal component regression (PCR) proposed by \cite{kendall1957course}, aims to establish a relationship between the input and the ground truth by identifying the multidimensional direction in the input space that best aligns with the maximum multidimensional variance direction in the output space. 

The second category comprises two straightforward neural network architectures that make pixel-based decisions. The first is the multilayer perceptron (MLP) \citep{PaolettiHPP19}, which is a simple feed-forward neural network with two layers. The second is the recurrent neural network (RNN) \citep{RumelhartHW86}, which utilizes recurrent connections to incorporate spectral information.

The following category combines basic convolutional neural networks with 1D, 2D and 3D kernels, allowing for the integration of spatial and spectral information. The 1D convolutional neural network (1D CNN) by \cite{PaolettiHPP19} employs 1D convolution layers only along the spectral dimension of each pixel. The 2D convolutional neural network (2D CNN) by \cite{PaolettiHPP19} convolves the input in the spatial dimension using 2D convolutional layers, while combining the spectral data in the fully connected head. And finally, the 3D convolutional neural network (3D CNN) by \cite{PaolettiHPP19} employs 3D convolutional layers that operate on all three dimensions of the hyperspectral cube. The difference in size of the models for the different approaches is obvious.

Furthermore, two additional configurations are evaluated using the 2D CNN architecture. The "spatial" configuration solely focuses on spatial features by reducing the $\lambda$ dimension to a single component, such as taking the mean of all channels. In contrast, the 'spectral' configuration exclusively utilizes spectral features from the center pixels without considering spatial context. This configuration utilizes the same information as the 1D CNN model.

The fourth category includes models that combine preprocessing filters with convolutional neural networks. One such model is the Gabor CNN \citep{GhamisiMLSTM18}, which applies a spatial Gabor filter to preprocess the hyperspectral input. The Gabor filter enhances textural information by exploring a higher-dimensional Gabor space. Another model in this category is the extended morphological profiles convolutional neural network (EMP-CNN) \citep{GhamisiMLSTM18}, which utilizes mathematical morphology operators such as opening and closing to standardize the input. Both of these methods are beneficial when textural information is crucial and have demonstrated successful performance on HRSS data sets. Based on the complexity of the filter calculation, these models are applied on a PCA-reduced input.

ResNet proposed by \cite{HeZRS16} is a widely used architecture commonly employed for color image data. Its key feature is the inclusion of skip connections, also known as shortcut connections, which enhance interconnectivity between layers at varying depths. This architecture is available in different layer configurations, and for our experiments, we utilize ResNet-18 with 18 layers and ResNet-152 with 152 layers. The latter represents a larger-scale CNN architecture.

In addition to these, modified versions of the ResNet networks are evaluated in our study. We replace the initial convolutional layers of the architectures with HyveConv layers, as proposed by \cite{VargaMBZ23}. This replacement not only reduces the number of trainable parameters to some extent, but more importantly, it ensures camera-agnostic models. This camera-agnostic capability is essential for training on recordings from different cameras, which will also be utilized for pretraining (Section \ref{sec:pretraining}).

The next category encompasses the DeepHS-Net architectures, which have been specifically optimized for ripeness classification of fruit using hyperspectral imaging (HSI). DeepHS-Net \citep{VargaMZ21} is a 2D convolutional neural network designed for efficient performance on small hyperspectral data sets. DeepHS-Hybrid-Net \citep{VargaFZ23}, on the other hand, combines 3D and 2D convolutions, making it a hybrid model. By leveraging both 3D and 2D convolutions, the network can reduce the number of parameters compared to a fully 3D CNN while still retaining the ability to convolve along the $\lambda$ dimension. Further, this category contains a DeepHS-Net with a HyveConv layer in the first layer, as proposed by \cite{VargaMBZ23}.

The next category includes two state-of-the-art methods specifically developed for the HRSS data set. One of these methods is SpectralNET \cite{Chakraborty2021SpectralNETES}, which utilizes wavelet transformations to conduct convolutions in both the spatial and spectral dimensions. The second method, HybridSN \citep{RoyKDC20}, is a hybrid architecture that combines 3D convolutions for spatial-spectral operations with 2D convolutions for purely spatial operations. It is worth noting that HybridSN has a significantly larger number of trainable parameters compared to DeepHS-Hybrid-Net. As a remark, the noted trainable parameters for HybridSN in Table \ref{tab:model_overview} are larger than in the original publication, as we increase the input patch size for all models (described in Sec. \ref{sec:training_procedure}). 

\begin{table*}[]
    \centering
    \def\arraystretch{1.5}
    \begin{tabular}{ll|rrr|rr}
    \hline
     &  & \textbf{HRSS} & \textbf{Fruit} & \textbf{Debris} & \textbf{Overall} & \textbf{Rank.}\\ \hline
    SVM &	 \multirow{2}{*}{Classical ML} &	78.85 \%&	47.10 \%&	52.57 \%&	59.50 \% & 16\\	
PLS-DA  &	 & 66.61 \%	 & 51.22 \%	 & 38.56 \%	& 52.13 \% & 21  \\	
\hdashline
MLP &	 \multirow{2}{*}{\begin{tabular}{@{}l@{}}
                Basic\\
                neural networks
         \end{tabular}}  &	71.18 \%&	44.54 \%&	50.72 \%&	55.48 \% & 22\\	
RNN  &	 &	69.61 \%&	41.72 \%&	52.50 \%&	54.61 \% & 23\\	
\hdashline
1D CNN  & \multirow{5}{*}{\begin{tabular}{@{}l@{}}Convolutional\\ neural\\ networks\end{tabular}}	 &	91.04 \%&	51.30 \%&	63.13 \%&	68.49 \% & 11\\	
2D CNN & &	99.71 \%&	54.42 \%&	77.39 \%&	77.17 \% & \underline{2}\\	
2D CNN  (spatial)& &	99.69 \% &	44.85 \% & 54.23 \%	 &	66.26 \% & 13 \\	
2D CNN  (spectral)&	 &	86.73 \%&	49.27 \%&	50.47 \%&	62.15 \% & 15 \\	
3D CNN  &	 &	\underline{99.73 \%}&	56.06 \%&	\textbf{87.56} \%&	\textbf{81.12} \% & \textbf{1}\\	
\hdashline
Gabor CNN &	 \multirow{2}{*}{CNNs + Filter} &	\textbf{99.75} \%&	52.57 \%&	66.43 \%&	72.92 \% & 5\\	
EMP CNN &	 &99.54 \% &	52.76 \% &	61.87 \% &	71.39 \% & 8 \\	
\hdashline
ResNet-18 &	 \multirow{4}{*}{ResNet networks} &	99.52 \%&	49.05 \%&	59.56 \%&	69.38 \% & 9\\	
ResNet-152 &	 &	96.27 \%&	47.00 \%&	29.30 \%&	57.52 \% & 19\\	
ResNet-18+HyveConv &	 &	99.67 \%&	51.43 \%&	67.12 \%&	72.74 \% & 6\\	
ResNet-152+HyveConv &	 &	97.22 \%&	42.66 \%&	46.91 \%&	62.26 \% & 18\\	
\hdashline
DeepHS-Net &	 \multirow{3}{*}{DeepHS networks} &	98.33 \%&	\textbf{58.28} \%&	75.32 \%&	77.31 \% & 4\\	
DeepHS-Net+HyveConv &	 &	98.53 \%&	\underline{57.57 \%}&	75.22 \%&	77.11 \% & 3\\	
DeepHS-Hybrid-Net &	 &	95.36 \%&	55.01 \%&	\underline{82.14} \%&	\underline{77.50} \% & 7\\	
\hdashline
SpectralNET  &	 \multirow{2}{*}{\begin{tabular}{@{}l@{}}State-of-the-art \\ for HRSS\end{tabular}} &	98.38 \%&	49.25 \%&	46.33 \%&	64.65 \% & 14\\	
HybridSN &	 &	97.54 \%&	48.74 \%&	73.85 \%&	73.38 \% & 10\\	
\hdashline
Attention-based CNN  &	 \multirow{3}{*}{\begin{tabular}{@{}l@{}}Attention-based \\ approaches\end{tabular}} &	89.93 \%&	44.88 \%&	50.18 \%&	61.66 \% & 20\\	
SpectralFormer  & &	96.25 \%&	41.71 \%&	53.24 \%&	63.74 \% & 17\\	
HiT  &	 &	98.47 \%&	48.16 \%&	59.23 \%&	68.62 \% & 12\\	
 
    \hline
    \end{tabular}
    \caption{Overall classification accuracy for the individual models and the three different data sets. In addition, the ranking based on the average ranking on the individual data sets is given. \textbf{Bold} highlights the best and \underline{underlined} the second-best accuracy per data set as well as overall.}
    \label{tab:overall_accs}
\end{table*}

The final category represents the latest trend in computer vision, which is attention-based methods, particularly in the form of vision transformers. Three models were selected in this category, optimized specifically for the HRSS data set. The first is an attention-based CNN proposed by \cite{LorenzoTMN20}, which employs an attention mechanism solely for the spectral dimension without utilizing spatial context. SpectralFormer, introduced by \cite{HongHYGZPC22}, is an early adopter of the vision transformer approach for hyperspectral recordings. It utilizes an attention-based method with spatial context, incorporating skip connections for a more flexible backbone. HiT, proposed by \cite{YangCLZ22}, is a vision transformer model that includes two key components: 3-D convolution projection modules and convolution permutators to capture subtle spatial-spectral discrepancies. It is important to note that the vision transformer methods have higher complexity compared to other approaches, and they are still in their early stages of development.

\begin{figure*}[ht]
    \centering
    \includegraphics[width=.73\textwidth]{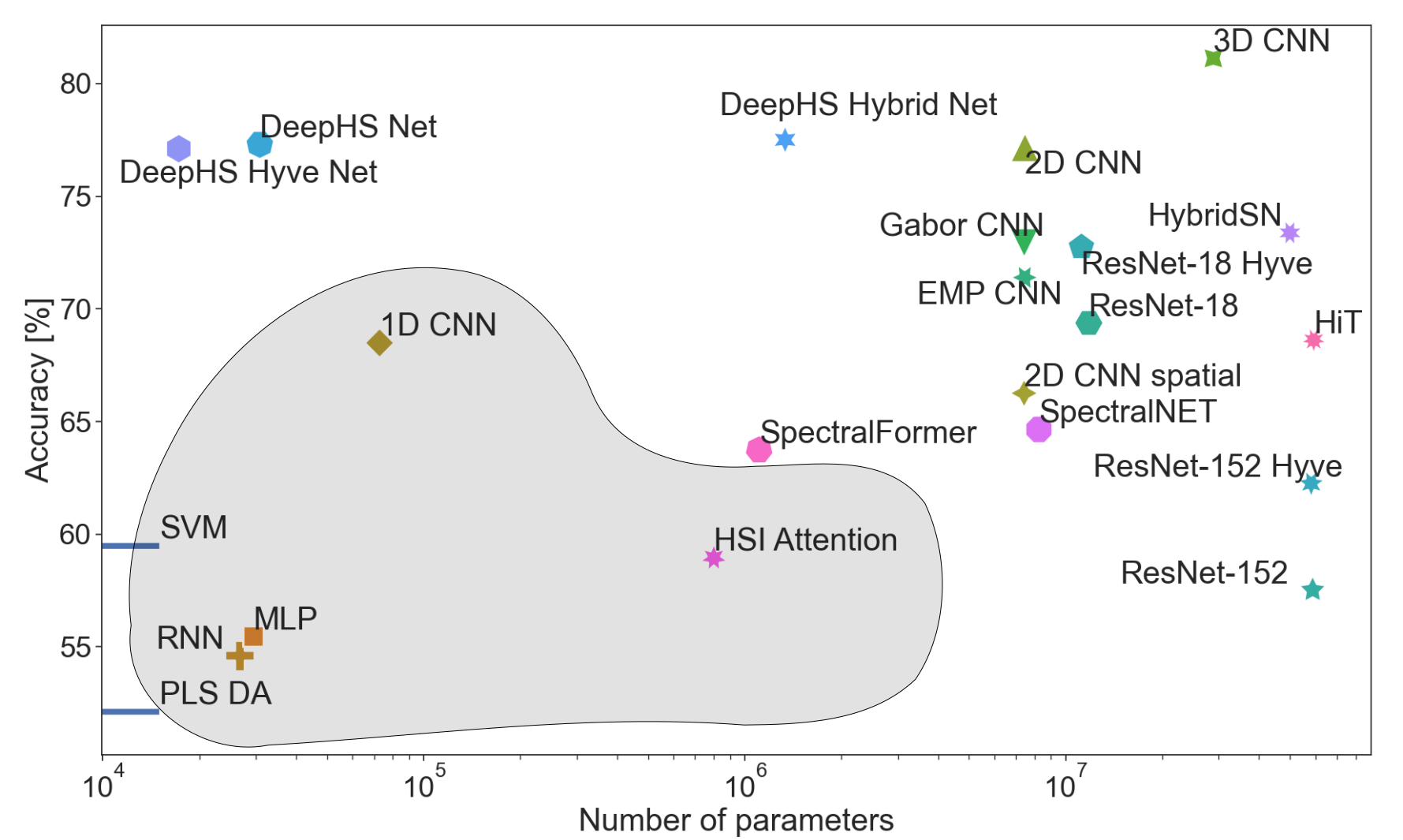}
    \caption{Overall classification accuracy versus size (number of parameters) of the individual models. The gray area contains all models which are not considering the spatial information.}
    \label{fig:modelsize}
\end{figure*}

In summary, a diverse range of models has been presented as baselines for the proposed benchmark. The models can be categorized based on their handling of the input data. The majority of models utilize the normalized hyperspectral cube as the input data. In some cases, a dimension reduction technique is applied to reduce the $\lambda$ dimension. Principal component analysis (PCA), which reduces dimensions based on the variance, is a commonly used method for this purpose and is recommended for several selected models (refer to Table \ref{tab:model_overview}).
 
\subsection{Training Procedure}\label{sec:training_procedure}
In order to enable a fair comparison between all of these models, we homogenized the training procedure as far as possible.

For each of the three benchmark data sets, a fixed train-val-test-split was used (see Sec. \ref{sec:data}), and the size of the classes in the categories was balanced, respectively.
Also, we used a standardized input image size across data sets and models. For objectwise classification, the whole image was resized to $128 \times 128$ pixels while for patchwise classification, we used patches of size $63$ pixels, in combination with a dilation of one for the HRSS data sets and 30 for the DeepHS Debris data set, respectively. For testing, all available pixels were used (dilation 1).

Using three different seeds each, we trained each classifier model for overall $36$ combinations of data set, classification task, and sensor.

In all cases, the model parameters were optimized with Adam \citep{Kingma14}, using a learning rate of $0.01$, which was stepwise decreased during training. Cross-entropy loss \citep{Cybenko99} was used as loss function. 
We trained for $50$ epochs and used checkpoint callback and early stopping based on the validation loss \citep{Prechelt98}.
A batch size of $32$ was chosen.
The training data was augmented using random flipping, random rotation and random cut, each with a probability of $50 \%$, and random cropping with $10 \%$ probability.
For individual model-specific exceptions, see Appendix, Tab. \ref{tab:model_hparams}.

\subsection{Results}\label{sec:results}

In this section, we delve into a comprehensive evaluation of the selected models, as outlined in Section \ref{sec:models}, in the context of the proposed benchmark elucidated in Section \ref{sec:data}. Moreover, we undertake an analysis of interesting aspects of hyperspectral image classification and highlight interesting insights.

Tab. \ref{tab:overall_accs} shows the overall classification accuracy for the individual models and the three different data sets. Highlighted are the top two accuracies for each of the data sets and overall, respectively. 

\begin{figure*}[ht]
    \centering
    \includegraphics[width=.7\textwidth]{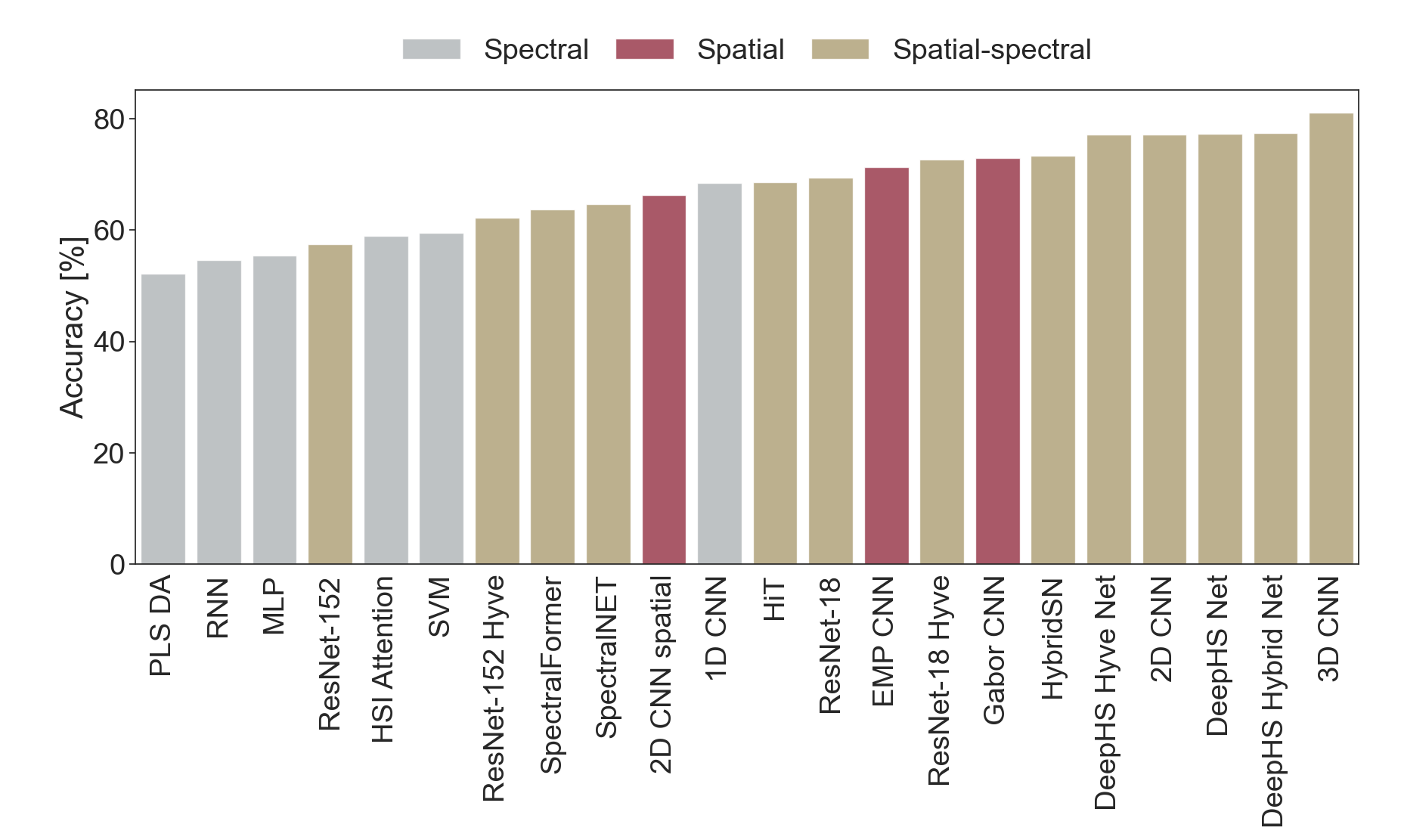}\label{fig:overall_spectralspatial}
    \caption{Overall classification accuracy for each model. Bars for models considering the spatial dimension in red, spectral in gray, both in gold.}
    \label{fig:spectral_spatial}
\end{figure*}

Classical ML methods, such as SVM, as well as simple neural networks, like MLP and RNN, perform rather badly over all data sets. 

The convolutional neural networks are optimized for the structure of image data. Among these, the 3D CNN emerges as the foremost performer, showcasing remarkable accuracy, particularly within the debris dataset. Notably, even the 2D CNN exhibits commendable performance. By exclusively feeding spatial or spectral data into the 2D CNN architecture, it becomes feasible to discern the significance of each feature in the decision-making process. For instance, in the context of the HRSS dataset, the spatial context emerges as a crucial determinant. A more intricate exploration of this phenomenon is undertaken in subsequent sections for an in-depth understanding.

Convolutional neural networks enhanced by specialized filters such as Gabor CNN or EMP CNN exhibit enhanced performance when applied to the HRSS dataset. However, despite this advantage, they fall short in effectively processing spectral features and consequently do not secure the top-performer position. Furthermore, their inference process tends to be computationally expensive.

The performance of deep convolutional neural networks, exemplified by the ResNet family, is relatively modest. Notably, the ResNet-152 network, while powerful, encounters challenges when dealing with diminutive hyperspectral datasets. Addressing this concern is a focal point of the subsequent section (Section \ref{sec:pretraining}), where pretraining methods come to the fore as a solution.

While the DeepHS-Net family was meticulously crafted to excel in the fruit dataset domain, delivering remarkable performance, its prowess on the HRSS dataset is more reserved. Nevertheless, it's important to emphasize that the overall performance remains consistently stable across various datasets.

The state-of-the-art models employed in our experiment exhibit commendable performance on the HRSS dataset. However, their performance falls short when applied to the other datasets. It is crucial to underscore that these models are evaluated with a same and fixed training-validation-test split for the first time.

The attention-based approaches did not show a significant improvement in comparison to the other models. Our experiments show that transformers are not necessarily better than the (state-of-the-art) CNNs for hyperspectral image classification. It seems that the global spectral information, which can only be processed by the transformer models, is not as important as the local context information captured by the convolutions.

\begin{figure*}
    \centering
    \subfigure[HRSS]{
    \includegraphics[width=.53\textwidth]{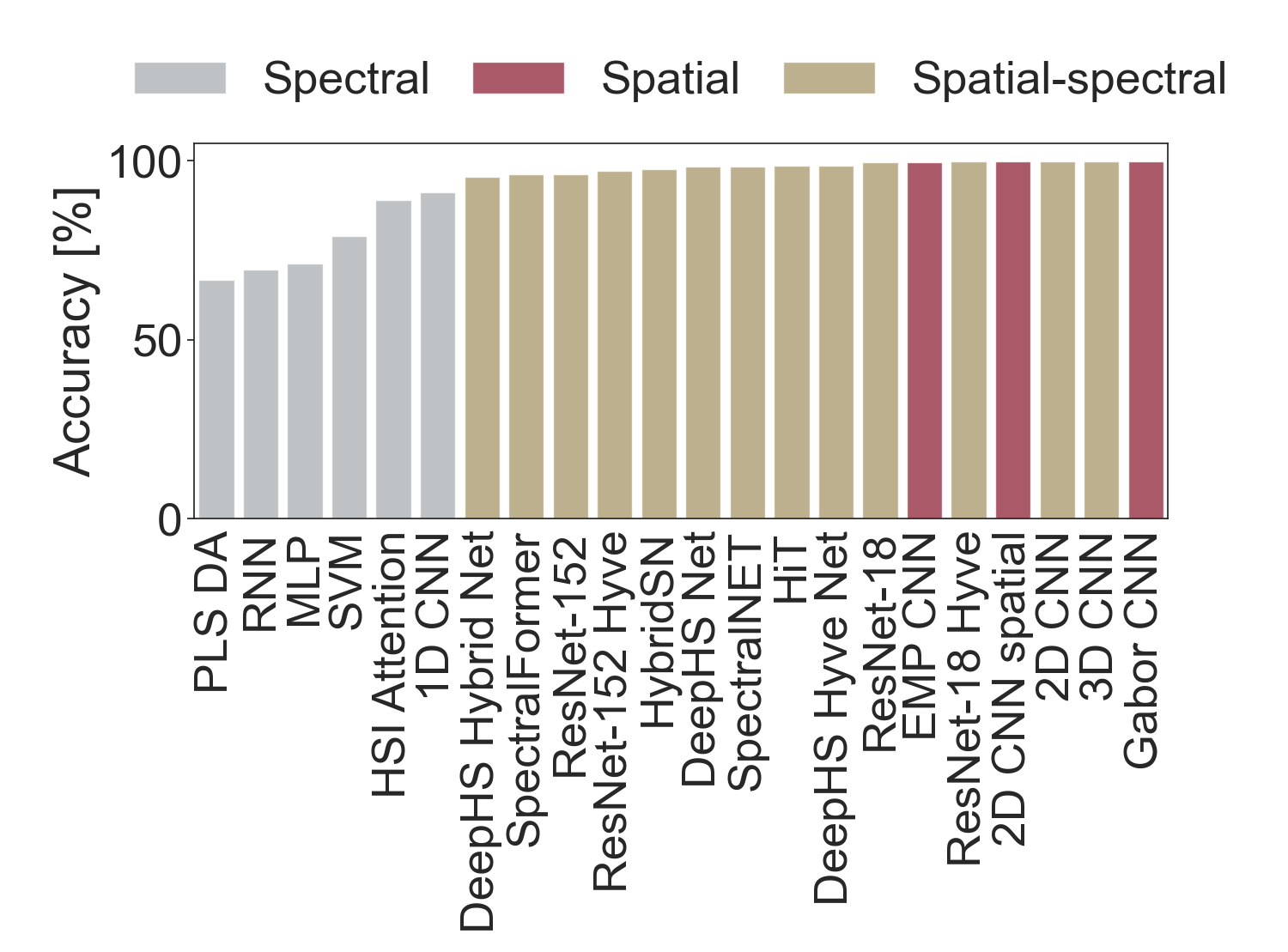}\label{fig:hrss_spatial}
    }
    \\
    \subfigure[DeepHS Fruit]{
    \includegraphics[width=.48\textwidth]{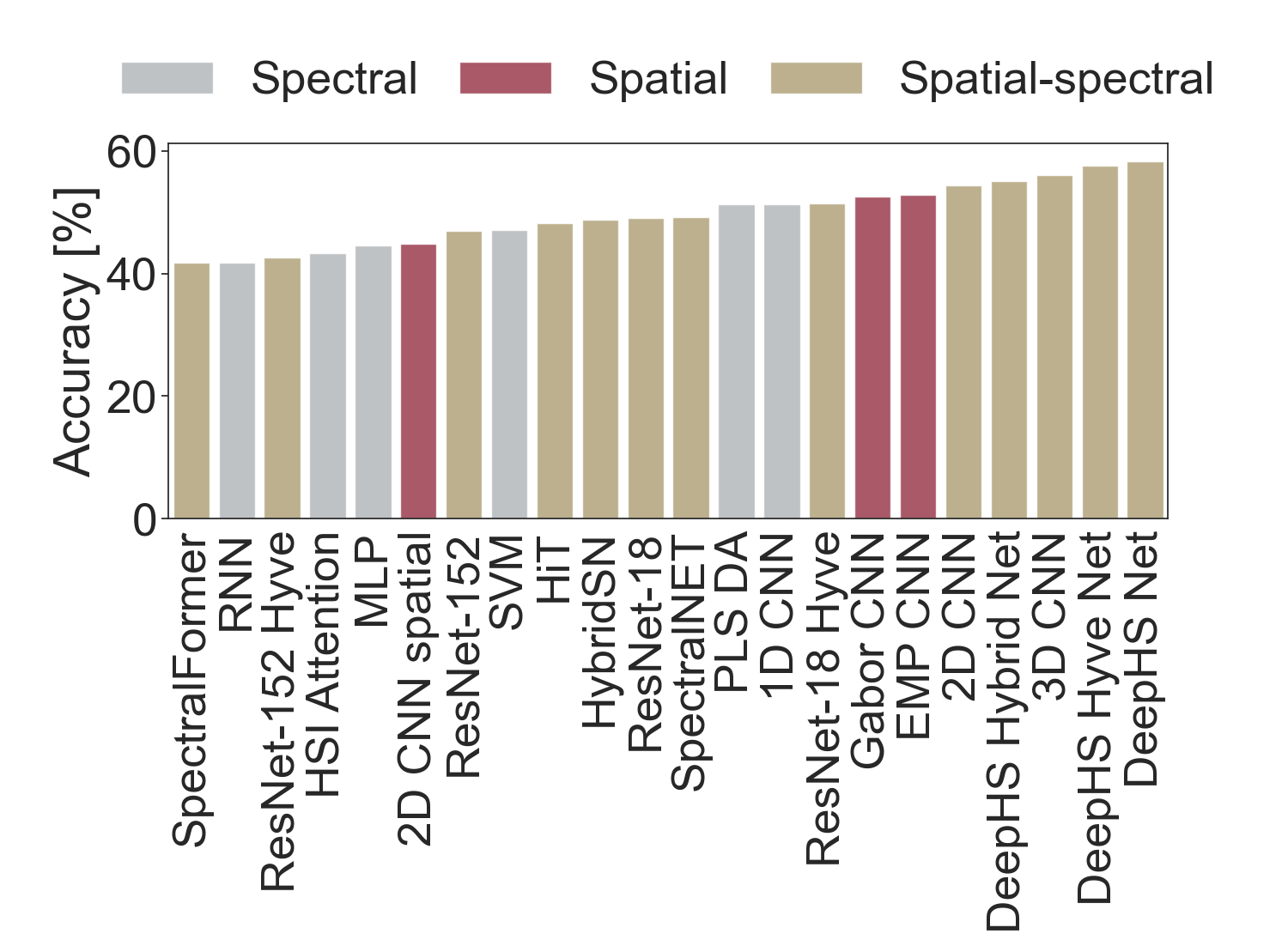}\label{fig:fruit_spectralspatial}
    }
    \subfigure[DeepHS Debris]{
    \includegraphics[width=.48\textwidth]{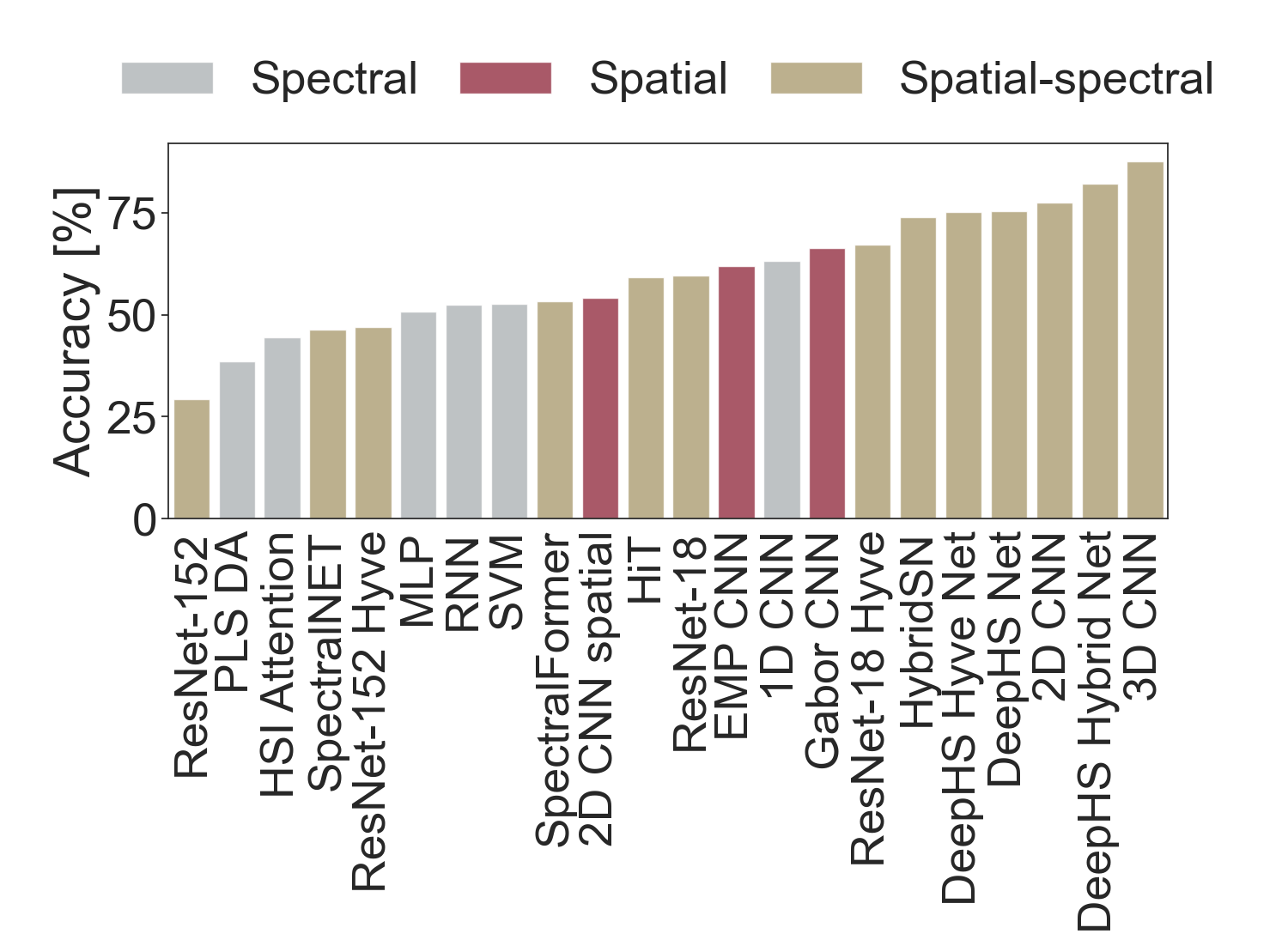}\label{fig:debris_spectralspatial}
    }
    \caption{Classification accuracy for the (a) HRSS data sets, (b) the DeepHS Fruit data set, and (c) the DeepHS Debris data set, for each model. Bars for models considering the spatial dimension in red, spectral in gray, both in gold.}
    \label{fig:individual_spectral_spatial}
\end{figure*}

To address the issue of varying accuracy variances among different datasets, we additionally introduce a model ranking in Table \ref{tab:overall_accs}. This ranking is established by computing the average placement of each model across individual datasets. Notably, a strong correlation between overall accuracy and the mean ranking emerges, albeit with a few outliers. This underscores that a mere comparison of overall accuracy is informative, but not sufficient. Consequently, an evaluation of performance across the individual datasets is addressed later in this section.

In Fig. \ref{fig:modelsize}, we augment the accuracy results with supplementary data concerning the model size. The 3D CNN remains the foremost performer, albeit noticeably larger in scale by comparison. The DeepHS-Net family continues to exhibit commendable performance, and notably, in contrast to the 3D CNN, commands significantly smaller dimensions (bearing in mind that the x-axis is logarithmically scaled). Additionally, it's evident that the inclusion of HyveConv contributes to an enhanced performance of the larger ResNet models, such as ResNet-18 and ResNet-152.

Differences in model performance can further be attributed to their respective feature extraction approach. As hyperspectral data contains two spatial dimensions as well as the spectral dimension for the model to consider, we distinguish between models with purely spatial, spectral or both, spatial-spectral feature extraction. 

The shaded region in Fig. \ref{fig:modelsize} accentuates all models that operate without incorporating spatial context. These models are devoid of the supplementary insights offered by the hyperspectral cube and consequently demonstrate inferior performance.

\begin{figure}
    \centering
    \includegraphics[width=.48\textwidth]{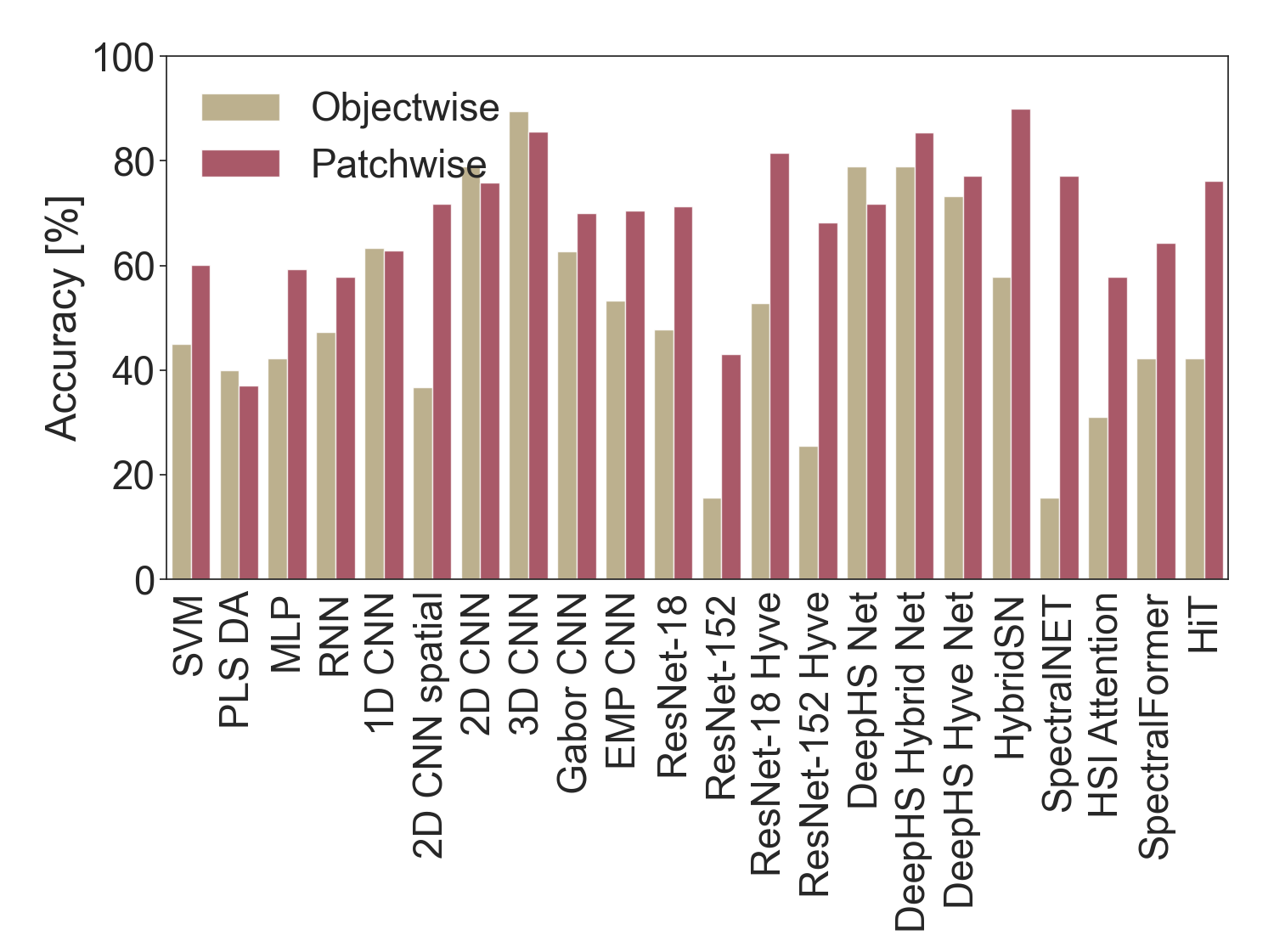}
    \caption{Classification accuracy for objectwise classification (gold) and patchwise classification (red) for the debris data set and for each model, respectively.}
    \label{fig:debris_objectpatchwise}
\end{figure}

Fig. \ref{fig:spectral_spatial} again shows the overall classification accuracy for all models. Bars corresponding to models considering the spatial dimension only are marked in red, spectral dimension only in gray, and both, spectral and spatial dimension, in gold.
Pixel-based models that only have access to the spectral information of a single pixel at a time achieve only low classification accuracies, while usage of the spatial information seems to be more relevant. 
Models like the DeepHS-Net variants or 3D CNN, operating on both, the spectral and spatial dimension, perform best overall.

Considering the classification performance on the remote sensing data exclusively (Fig. \ref{fig:hrss_spatial}), we obtain an even more extreme distribution.
All models that incorporate only the spectral dimension (e.g., MLP, RNN, 1D CNN) perform significantly worse than those including the spatial component.
We even find that, in this case, the purely spatial information is already enough to achieve $> 95\%$ mean accuracy. The 2D CNN (spatial), that operates on a single aggregated channel dimension, as well as the Gabor CNN are among the top five models, leading to the conclusion that the hyperspectral information contained in the channels is not needed here.


This should be alarming to any researcher working in the field, as it would indicate that the most popular and almost exclusively used hyperspectral data set is not so well suited for evaluating spectrum-based methods after all.
With our benchmark collection, we aim to provide alternative data sets and classification tasks,
for which we show that, as expected, models considering both the spectral and spatial information can actually outperform purely spatial models for hyperspectral image classification (see Fig. \ref{fig:fruit_spectralspatial} and \ref{fig:debris_spectralspatial}).

Further, we find that for at least some of the models, performance differs significantly for whether the task is objectwise or patchwise classification (see Fig. \ref{fig:debris_objectpatchwise}).
Especially large and complex models, like the attention-based models and the larger ResNet (i.e., ResNet-152), but also the models specifically designed for HSI classification, perform much worse for objectwise than for patchwise classification.

On the data side, the biggest difference is the amount of training samples available; every recording yields only a single sample with the objectwise approach, but on the other hand, can be divided into many patches, yielding many samples per recording for patchwise classification. For the DeepHS Debris data set, for example, we obtain $85$ versus $11,635$ training samples, respectively.
The extremely low number of training data samples might not be sufficient to optimize the large number of parameters of some of the models, and therefore explain their rather bad performance on the objectwise classification task.

Another explanation might be that the state-of-the-art HSI classifiers (e.g., HybridSN, SpectralNET, SpectralFormer) are optimized for the almost exclusively used remote sensing application and the corresponding segmentation task, and therefore on patchwise image processing.

In turn, the rather small models of the DeepHS-Net family were explicitly designed for handling small hyperspectral data sets and optimized for predicting fruit ripeness, which is an objectwise classification task, and could explain their outstanding performance on objectwise classification, also for the Debris data set.

\begin{figure*}
    \centering
    \includegraphics[width=.8\textwidth]{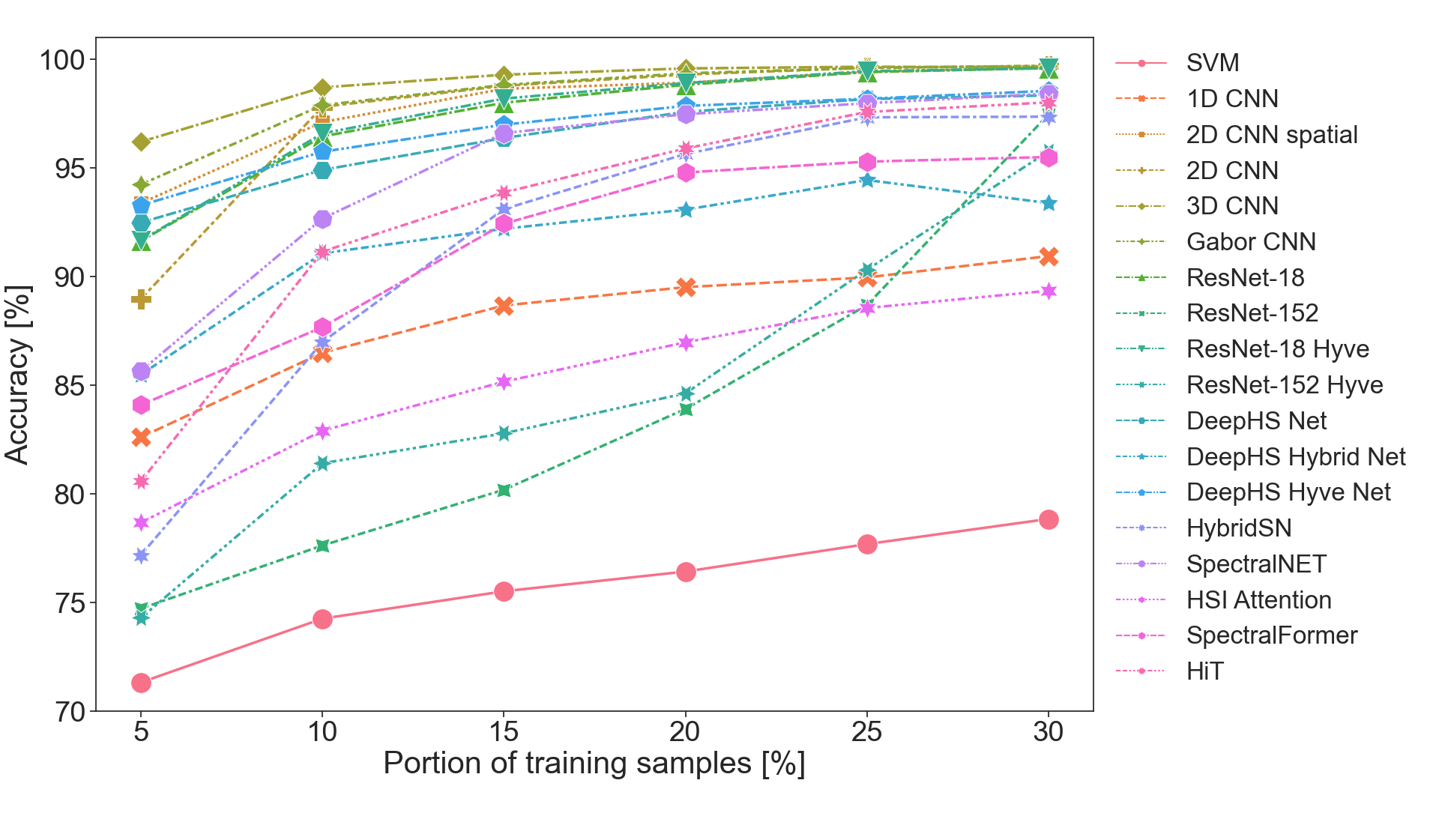}
    \caption{Classification accuracy versus portion of training samples (5 \%, 10 \%, 15 \%, 20 \%, 25 \% and 30 \%) for the HRSS data set and each of the models, respectively.}
    \label{fig:hrss_trainratio}
\end{figure*}

Motivated by the abovementioned observations, we further investigated the dependence of the model performance on the number of labeled training samples available. For the three HRSS data sets, Fig. \ref{fig:hrss_trainratio} shows the average classification accuracy as a function of the ratio of the labeled training data, which was stepwise reduced from 30\% to 5\%.

As to be expected, the general trend is that better performance is achieved with more training data.
However, on the basis of the models' concrete behavior, different groupings emerge:

The simple methods (e.g., SVM, PLS-DA, as well as MLP, and RNN) have low accuracies and also do not to get much better, even when provided with more training samples. In the plot, we included only the SVM in as a representative example, but disregarded the other three, as -- due to their below-average performance -- they are considered irrelevant in this context and will not be discussed further.

Further, we find a number of models in the middle field, achieving average accuracies while still following the trend of improving for larger portions of training data.

The smaller DeepHS-Net model variants, as well as the 3D CNN and 2D CNN are already quite good for only 5\% of the data, improving little when more training data is added, while still remaining among the best performing models (also compare to Tab. \ref{tab:overall_accs} and Fig. \ref{fig:modelsize}). For these models, the training data ratio seems not important -- they can also handle small data sets rather well.

But then, there is also a group of "data-hungry" models that show poor performance on small fractions of the training data, but whose classification accuracy rises significantly as more samples are provided. An outstanding example is ResNet-152, for which the accuracy increases almost linearly with the portion of labeled training samples. Similar behavior can be observed for more complex hyperspectral classifiers (e.g., HybridSN) and some transformer models. 
The most probable explanation for this is, once again, based on the complexity and size of the models. The latter need more training data to optimize their large number of parameters, and therefore can only show satisfying performance when enough training samples are available.

This immediately brings up the question whether the larger models, like the ResNet-152, could become much better and even as good as, e.g., the DeepHS-Net (+ HyveConv) model, if they only were to see more training data.

This is addressed in Section \ref{sec:pretraining} where we provide and evaluate a strategy for pretraining the model on other, yet similar hyperspectral image data.

\section{Pretraining and Transfer Learning}\label{sec:pretraining}

\begin{figure*}
    \centering
    \includegraphics[width=.75\textwidth]{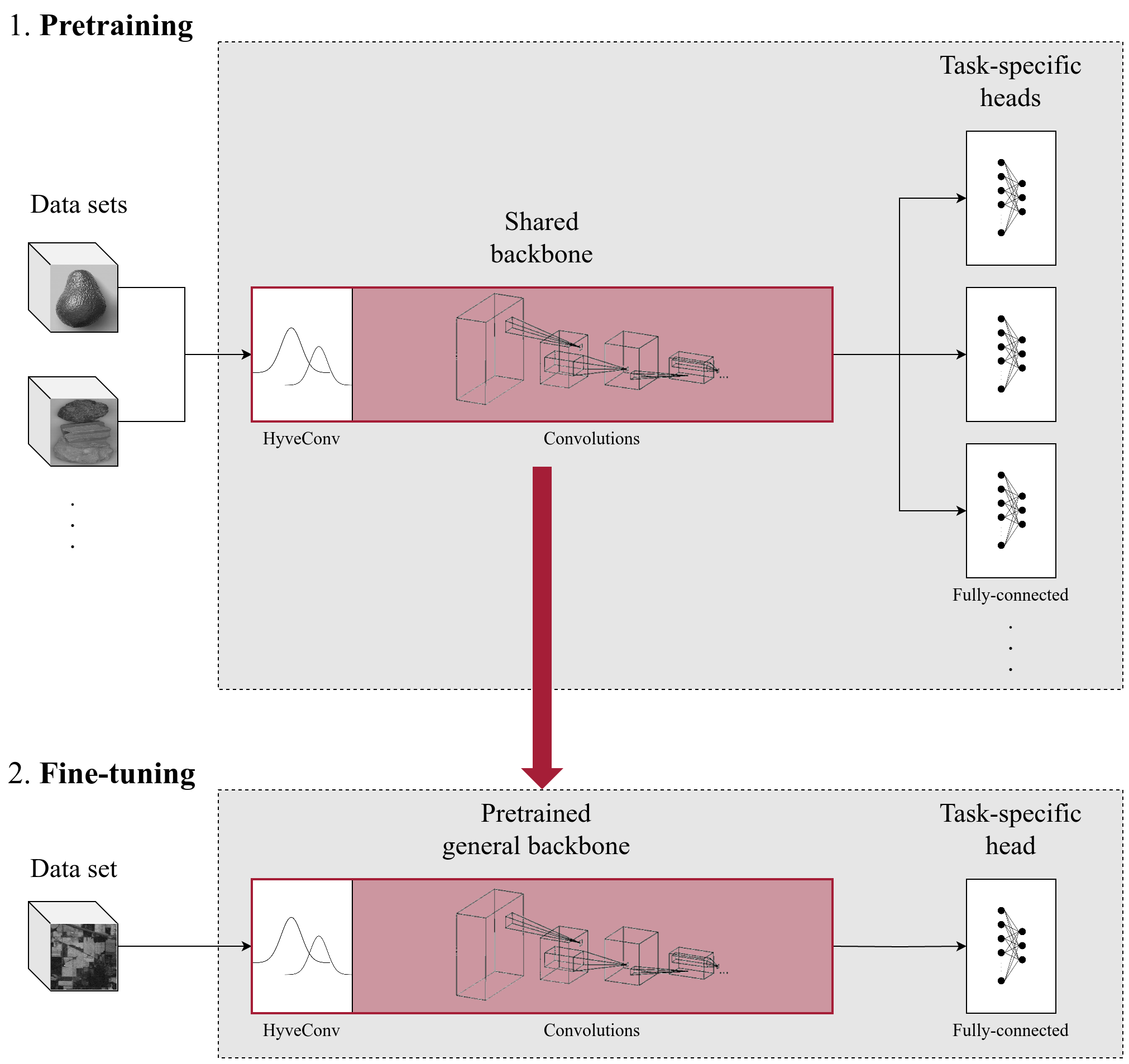}
    \caption{Scheme of the proposed pretraining and fine-tuning procedure. 
    First, we pretrained the model on (potentially) multiple data set configurations (i.e., data sets, cameras and classification tasks), using a shared backbone with an initial HyveConv layer and multiple task-specific heads.
    We kept the general pretrained backbone weights, while re-initializing the task-specific fully-connected part for fine-tuning on a single specific data set and task.
    }
    \label{fig:pretraining_scheme}
\end{figure*}

As we have found, some of the classifier models highly depend on the amount of training data available (see Fig. \ref{fig:hrss_trainratio}). 
However, labeled hyperspectral data is usually rather scarce, 
and even for the simplest networks, training solely on one small data set, can lead to unstable training and overfitting.

As an effective workaround, 
we propose 
pretraining the model on other, potentially multiple hyperspectral data sets and tasks collected in this benchmark beforehand,
and then just fine-tuning it using the data and configuration for the actual classification task
-- to improve classification accuracy, avoid overfitting and consequently allow using larger (deeper) models in general.

This idea was successfully applied for large RGB data sets (e.g., \cite{KrizhevskySH12}). 
Also, pretraining for hyperspectral image classification has already been explored to some extent, e.g., by \cite{LeeEK19, LeeEK22, WindrimMMCR18}. \cite{LeeEK22} showed that it also works in this special case, and that it is possible to pretrain a shared backbone using a so-called multi-domain approach.
However, they only considered different remote sensing scenes, with both, data and task, still being very similar across those domains.

In contrast, utilizing our benchmark data sets, we look at entirely different application scenarios where data, recorded wavelengths and classification task vary considerably.
We address the following questions: 
Is it still reasonable and helpful to combine the different data sets and train a shared backbone on multiple tasks?
Can we even benefit from this large variety to find a common structure in HSI data and extract the most general hyperspectral features -- independent of application, task, and sensor?
And finally, can we transfer these learned features to an unseen target data set and task in the fine-tuning process?

\subsection{Pretraining Strategy}\label{sec:pretraining_strategy}
To answer these questions, in this section, we present a pretraining and fine-tuning strategy to pretrain any model on potentially multiple different hyperspectral data sets and tasks, and subsequently fine-tune and evaluate it on a specific target configuration.

We assume a HSI classifier model of most general structure,
consisting of a backbone for feature extraction and a task-specific head.
The backbone is usually built out of multiple convolutional layers, including a first layer which depends on the data set (wavelengths; number of channels) and the remaining part of the backbone, which is more abstract and independent of the specific data set and task.
The head, one or few fully-connected linear layer(s), is again specific to the task at hand and carries out the actual classification, and therefore depends on the number of output classes.

Two major adjustments were made for our purposes:
\begin{itemize}
\item 
To allow training on multiple data sets as well as to adapt to a differing target data set easily, the first convolutional layer was replaced by a hyperspectral visual embedding convolution (HyveConv) layer \citep{VargaMBZ23}. It operates on a wavelength-based feature learning paradigm, rather than the conventional channel-based approach and therefore ensures camera-agnostic models, applicable across different camera setups.
The HyveConv was used in combination with an extended wavelength range, covering the lowest and highest wavelength of all data under consideration, 
avoiding the need to employ a separate first layer for different sensors.
\vspace{0.1cm}
\item 
Further, we introduced a multi-head approach similar to \cite{Lee18} to pretrain on multiple tasks, with (potentially) different class outputs, simultaneously. For each data set and classification task, another fully-connected head was used.
We balanced the data sets and used mixed batches (equal ratio per configuration and batch).
We switched between heads, depending on the current task.
As suggested by \cite{LeeEK19, LeeEK22}, the learning rate was multiplied by a factor $\frac{1}{N}$, when $N$ different configurations were considered simultaneously.
\end{itemize}

As visualized in Fig. \ref{fig:pretraining_scheme}, we then employed the following general pretraining and fine-tuning procedure.

\begin{enumerate}
    \item \textbf{Pretraining.}
    For pretraining on (potentially) multiple data sets, cameras and / or classification tasks, 
    as described above, we used the HyveConv layer, and switched between multiple task-specific heads, depending on the current data set configuration. The training was conducted on mixed batches and with a reduced learning rate (see above).
    \vspace{0.2cm}
    \item \textbf{Fine-tuning.}
    To specialize on a specific data set, camera and classification task, 
    first, we reinitialized the fully-connected task-specific head (except for the BN layer) to adapt the class outputs, while keeping the pretrained weights of the remaining intermediate layers, the "shared backbone". 
    Then, the model was again optimized in an end-to-end fashion,
    effectively training the last layers from scratch, while only fine-tuning the general backbone part.
\end{enumerate}

\subsection{Experiments}
For the pretraining experiments, we constrained ourselves to two kinds of models, variants of the DeepHS-Net \cite{VargaMZ21, VargaMBZ23} and the ResNet \cite{HeZRS16}.
By modifying them as described in Sec. \ref{sec:pretraining_strategy}, we obtain a multi-head DeepHS-Net+HyveConv, as well as a multi-head ResNet-18+HyveConv, and ResNet-152+HyveConv. In addition, we extended the multi-head DeepHS-Net+HyveConv to obtain a larger variant of this model (5 instead of 3 hidden layers of increased size; parameters), mainly for comparison.

We pretrained the respective multi-head model on a subset of our benchmark data, using an initial learning rate of $\frac{0.01}{N}$ for $N$ different data set configurations, which was stepwise decreased during training. Cross-entropy loss \citep{Cybenko99} was used as loss function, optimized using Adam \citep{Kingma14}. We trained for $50$ epochs and used checkpoint callback and early stopping based on the validation loss \citep{Prechelt98}. In each epoch, mixed batches of size $32$, were considered. As for regular classification (see Sec. \ref{sec:training_procedure}), the pretraining data was augmented using random flipping, random rotation and random cut, each with a probability of $50 \%$, and random cropping with $10 \%$ probability.

To fine-tune on the given data set, camera and task,
after (re-)initializing the fully-connected task-specific head -- except for the BN layer --,
the model was again trained as described in Sec. \ref{sec:training_procedure}, for $50$ epochs with checkpoint callback and early stopping, using an initial learning rate of $0.01$ with a stepwise-decrease, the cross-entropy loss, Adam optimizer, and a batch size of $32$. The same set of data augmentations was applied.

Again, we conducted the experiments for 3 different seeds each and report the average classification accuracy as well as their standard deviation.


We tested the proposed pretraining strategy on two hyperspectral remote sensing scenes data sets and the corresponding patchwise classification task.
Each of the abovementioned models was pretrained on HRSS/Salinas with a train ratio of 30\%. For fine-tuning and evaluation on HRSS/Indian Pines, we deliberately choose to consider only 5 \% of the pixels to obtain more meaningful results. Otherwise, both accuracy values -- without and with pretraining -- were near 100 \% already, and therefore hard to compare.
Tab. \ref{tab:pretraining_} lists the classification accuracies obtained with and without pretraining, respectively.

\begin{table*}[]
    \centering
    \def\arraystretch{1.2}
\begin{tabular}{crrrr}
\hline
               & \multicolumn{1}{c}{\begin{tabular}[c]{@{}c@{}}DeepHS-Net\\ +HyveConv\end{tabular}} & \multicolumn{1}{c}{\begin{tabular}[c]{@{}c@{}}Larger\\ DeepHS-Net\\ +HyveConv\end{tabular}} & \multicolumn{1}{c}{\begin{tabular}[c]{@{}c@{}}ResNet-18\\ +HyveConv\end{tabular}} & \multicolumn{1}{c}{\begin{tabular}[c]{@{}c@{}}ResNet-152\\ +HyveConv\end{tabular}} \\ \hline
No pretraining & 81.42 \%                                                                           &       79.19 \%                                                                                                                        & 77.64 \%                                                                          & 30.48 \%                                                                           \\
Pretraining    & \textbf{88.19 \%}                                                                  & \textbf{89.19 \%}                                                                                                           & \textbf{92.76 \%}                                                                 & \textbf{90.91 \%}                                                                  \\ \hline
\end{tabular}
    \caption{Classification accuracy (mean and standard deviation) with pretraining and without pretraining, for different models (DeepHS-Net+HyveConv, the larger DeepHS-Net+HyveConv, ResNet-18+HyveConv, and the ResNet-152 +HyveConv). The higher accuracy is marked in \textbf{bold}, respectively.}
    \label{tab:pretraining_}
\end{table*}

In  all cases, the pretrained model performs better than without pretraining, respectively. Pretraining on another remote sensing scene boosts classification accuracy by a significant amount of $6 \%$ up to over $60 \%$.
We observe a strong correlation between the model size and this performance gain; the larger the model, the more of an impact pretraining has and the more of an improvement it brings.
Already when comparing the DeepHS-Net+HyveConv versus its larger version, the gain in accuracy is almost doubled.
Further, the improvement is larger for the ResNet-18+HyveConv, and even larger ($60.43\%$) for the ResNet-152+HyveConv model.
All the three larger models that initially performed worse, achieve an even higher classification accuracy than the DeepHS-Net+HyveConv model, when pretrained.

\begin{figure}
    \centering
    \subfigure[]{\includegraphics[width=.35\textwidth]{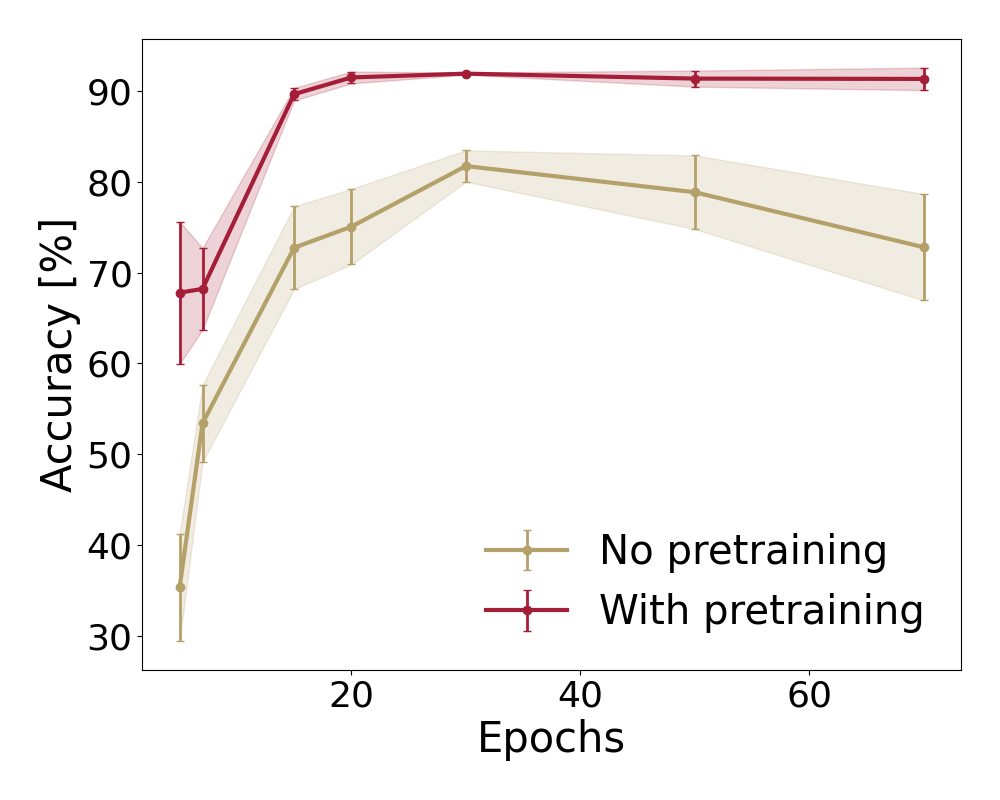}\label{fig:pretraining_epochs}}
    \subfigure[]{\includegraphics[width=.35\textwidth]{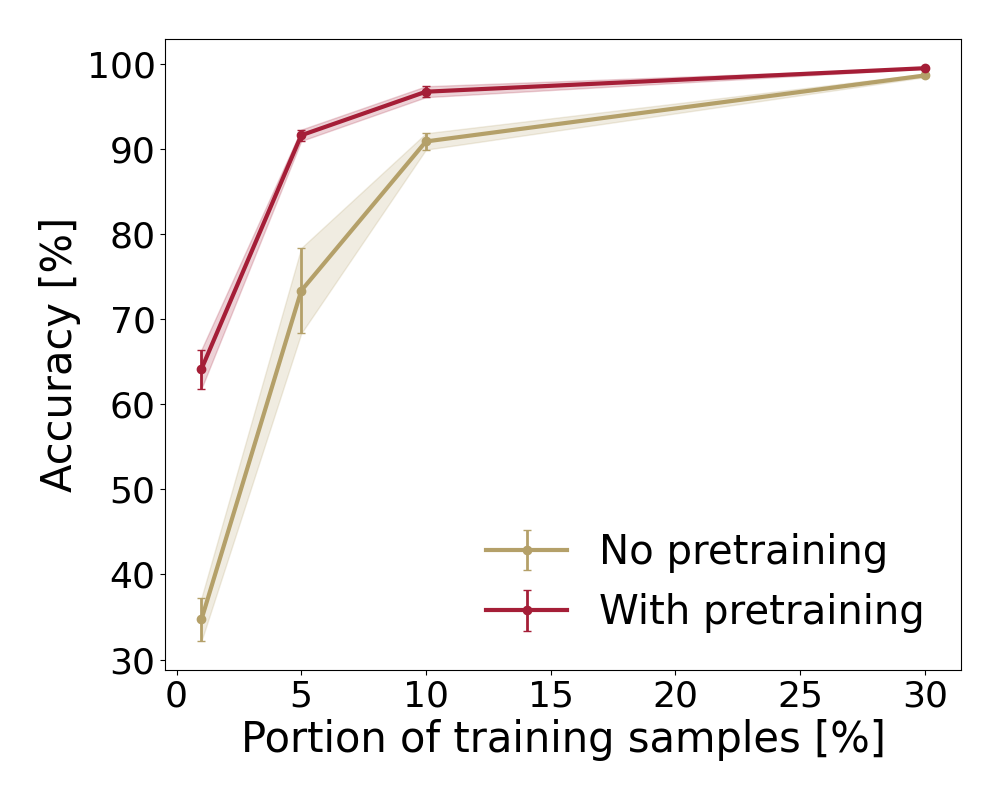}\label{fig:pretraining_trainratio}}
    \caption{Classification accuracy (mean and standard deviation) versus (a) the number of fine-tuning epochs and (b) the portion of training samples, for the ResNet-18+HyveConv model. The curve for the pretrained model is marked in red and for the model without pretraining in gold, respectively.}
    \label{fig:pretraining}
\end{figure}

As usually, the advantage of pretrained models comes most to effect when very few labeled training samples are available and / or subsequent fine-tuning is limited (e.g., by time or resources), we also examine the dependence of the classification performance on the portion of training samples ($1 \%$ up to $30 \%$, see Fig. \ref{fig:pretraining_trainratio}) as well as on the number of epochs for fine-tuning (Fig. \ref{fig:pretraining_epochs}). 

As expected, in both cases -- without and without pretraining --, the classification accuracy increases with increasing ratio of training samples used.
However, with pretraining, the accuracy is already quite high for only a small fraction of the training data, and the corresponding curve remains above the curve without pretraining for all the remaining ratios measured, indicating a higher overall accuracy with pretraining.

Analogously, with pretraining, we observe an already high accuracy for low number of training epochs, and again, a higher accuracy overall.
Without pretraining, the classification performance even decreases again after a certain number of epochs, possibly due to overfitting when no pretraining is employed beforehand.

We conclude that, with pretraining, in general, fewer epochs and less labeled training data samples were needed for the subsequent training on the target data set and task.

For all (other) combinations of pretraining and fine-tuning on all data sets and tasks included in our proposed benchmark, Tab. \ref{tab:pretraining_comb} shows the improvement in accuracy with pretraining relative to pure classification, 
using one representative example configuration, but covering the whole spectrum of applications, sensors, and tasks, respectively:
The Salinas remote sensing data set (with a train ratio of 30\%), the avocado fruit measurements recorded by the Specim FX10 camera w.r.t. firmness, and the debris data recorded by the Corning HSI -- patchwise and objectwise, respectively -- were used for pretraining, while for fine-tuning and evaluation, we considered the following configurations: HRSS Indian Pines (with only 5\% train ratio), the avocado data recorded by the Specim FX10 -- this time w.r.t. ripeness, and the debris patchwise / objectwise data, recorded also by the Specim FX10.

\begin{table*}[]
    \centering
    \def\arraystretch{1.5}
\begin{tabular}{cl|>{\raggedleft\arraybackslash}p{2.1cm}>{\raggedleft\arraybackslash}p{2.1cm}>{\raggedleft\arraybackslash}p{2.1cm}>{\raggedleft\arraybackslash}p{2.1cm}}
\hline
\multicolumn{1}{l}{}                  &                   & \multicolumn{4}{c}{\textbf{Fine-tuning and evaluation}}                                                                             \\
\multicolumn{1}{l}{}                  &                   & \multicolumn{1}{c}{HRSS} & \multicolumn{1}{c}{Fruit} & \multicolumn{1}{c}{Debris (patchw.)} & \multicolumn{1}{c}{Debris (objectw.)} \\ \hline
\multirow{4}{*}{\textbf{Pretrain.}} & HRSS              & \textbf{+ 15.12 \%}               & + 2.78 \%                 & + 4.54 \%                            & + 10.00 \%                            \\
                                      & Fruit             & + 3.44 \%                & + 9.73 \%                 & + 9.36 \%                            & $\pm$ 0.00 \%                               \\
                                      & Debris (patchw.)  & + 13.58 \%               & \textbf{+ 18.06 \%}                & \textbf{+ 15.17 \%}                           & \textbf{+ 33.33 \%}                            \\
                                      & Debris (objectw.) & + 7.00 \%                & + 5.56 \%                 & + 6.79 \%                            & + 3.33 \%                             \\ \hline
\end{tabular}
    \caption{Average improvement in classification accuracy with pretraining relative to pure classification without pretraining, exemplary for all combinations of data set and task, for pretraining and fine-tuning, and the ResNet-18+HyveConv model.}
    \label{tab:pretraining_comb}
\end{table*}

Foremost, in all cases except for one, the classification performance could be improved when pretraining on either one of the four example configurations.

It is worth emphasizing that pretraining on the patchwise classification task and the debris data set could increase the accuracy by over $10 \%$ for all the three data sets, different sensors and classification types.

Analyzing the results in more detail,
we also find that there is a significant improvement when pretraining and fine-tuning on (other) remote sensing scenes, probably due to the fact that those three data sets are very similar in terms of sensors and classes and share the same patchwise classification task, which makes feature transfer especially easy.
Similarly, for correlated features, like firmness and ripeness of a fruit, the transfer is possible.
However, considering the improvement for the fruit data set as well as the objectwise configuration for the debris data set, it seems that one cannot learn much from the few training samples provided for objectwise classification in general, 
while patchwise pretraining and fine-tuning on the objectwise classification task works surprisingly well. The outstanding example is again patchwise pretraining and objectwise fine-tuning for the debris data set, where we observe an improvement of $33 \%$ in classification accuracy,
although different sensors were used.
On the other hand, it also seems to help, if the data was recorded by the same sensor,
and feature transfer is possible for the same sensor, but differing tasks, for example.

These are promising results, since they show that it is indeed possible to transfer hyperspectral features between different data sets, tasks, and even classification types. Pretraining on hyperspectral data of all kind helps to increase classification performance.

In contrast, we show that pretraining on regular image data (i.e., RGB data) does not work.
Tab. \ref{tab:pretraining_RGB_HSI} lists the classification accuracy for two ResNet's of different size -- without pretraining, when pretrained on the RGB ImageNet data set \cite{Russakovsky14}, and with pretraining on hyperspectral data of two different kinds.

\begin{table}[]
    \centering
    \def\arraystretch{1.2}
\begin{tabular}{lrr}
\hline
 & \multicolumn{1}{c}{\begin{tabular}[c]{@{}c@{}}ResNet-18\\ +HyveConv\end{tabular}} & \multicolumn{1}{c}{\begin{tabular}[c]{@{}c@{}}ResNet-152\\ +HyveConv\end{tabular}} \\ \hline
No pretraining         & 77.64 \% & 30.48 \%  \\ \hline
Pretraining on ImageNet& 50.83 \% & 25.81 \%  \\ \hline
Pretraining on HRSS    & 92.76 \% & 90.91 \%  \\
Pretraining on Debris  & 91.22 \% & 85.21 \%  \\ \hline
\end{tabular}
    \caption{Mean classification accuracy for the ResNet-18+HyveConv and ResNet-152+HyveConv on HRSS Indian Pines (train ratio 5 \%) without pretraining and with pretraining on ImageNet and on two different other hyperspectral data sets (HRSS Salinas (train ratio 30 \%) and debris (Corning HSI, patchwise classification)).}
    \label{tab:pretraining_RGB_HSI}
\end{table}

Pretraining on ImageNet even decreases the accuracy relative to no pretraining while pretraining on the hyperspectral data increases the accuracy, even for an entirely different data set, leading to the conclusion that pretraining and feature transfer does only work for (other) hyperspectral data. 
This in turn strengthens the assumption that specialized methods are needed for this kind of learning on HSI data -- as apparently, pretraining on HSI data is more complex and not as easy and straight-forward as for regular RGB image data (see, e.g., \cite{LeeEK19}).

We were able to prove that our specialized approach to pretraining on hyperspectral data works, not only for a single pretraining configuration, but also when combining multiple hyperspectral data sets, sensors and tasks for so-called multi-task pretraining. In this sense, we could make the best possible use of the variety of data and classification tasks in the proposed benchmark, 
with the goal to learn more general features for HS image classification, independent of the concrete application.



\begin{figure*}[]
    \centering
    \includegraphics[width=0.70\textwidth]{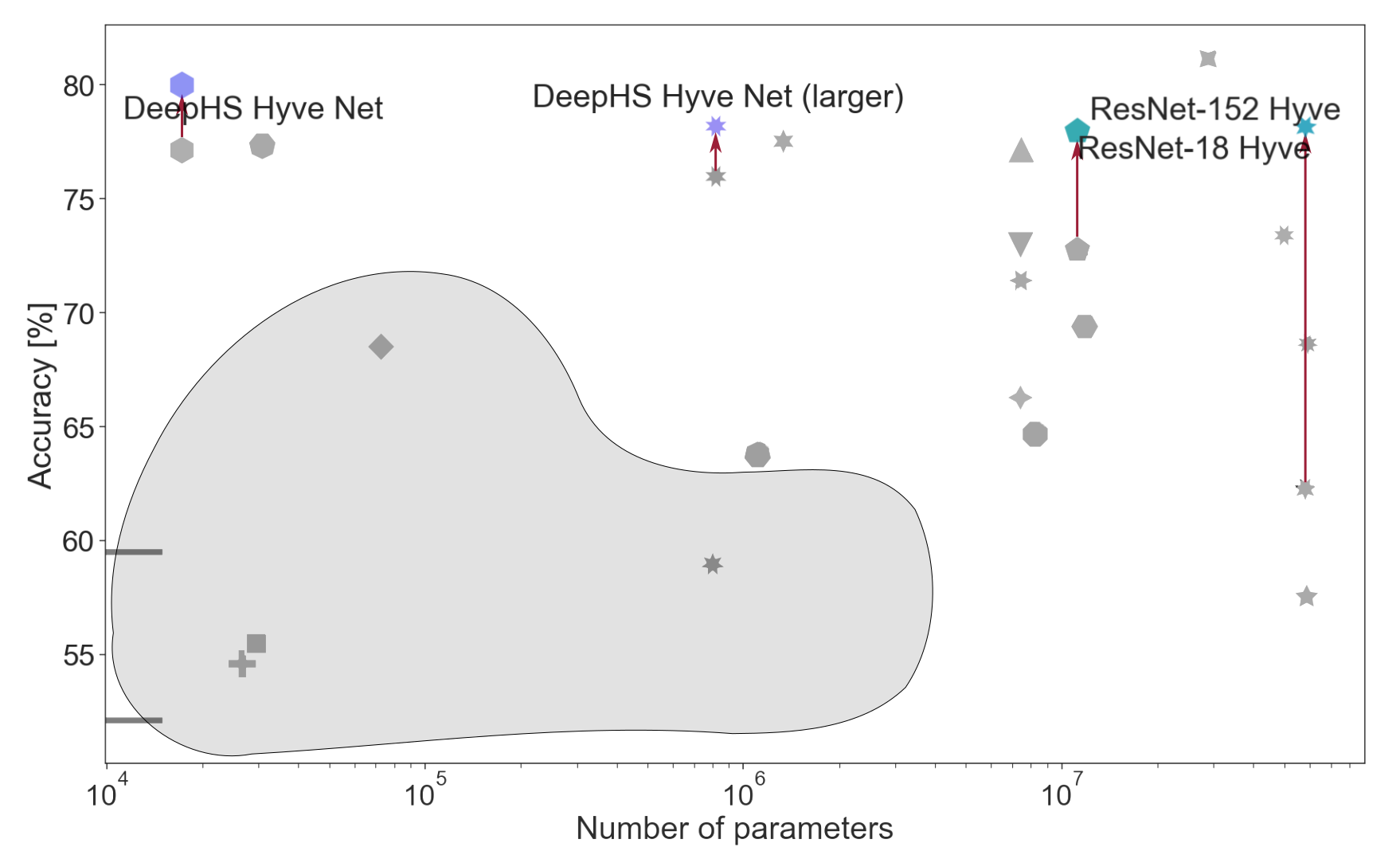}
    \caption{Increase in overall classification accuracy without and with pretraining in relation to the model size, for the DeepHS-Net+HyveConv, larger DeepHS-Net+HyveConv, ResNet-18+HyveConv, and ResNet-152+HyveConv, respectively.}
    \label{fig:overall_wo_pretraining}
\end{figure*}

We pretrained the abovementioned models on a selection of different configurations, covering almost all data sets, sensors, and classification tasks, and evaluated them again on all possible configurations in the benchmark (see Appendix, Tab. \ref{tab:training_val_test_splits}), reporting the overall classification accuracy without and with pretraining, respectively.

Fig. \ref{fig:overall_wo_pretraining}, similar to Fig. \ref{fig:modelsize}, shows those accuracy values in relation to the model's size, with the pretrained versions of the models of interest added to the plot.

Again, we observe that, in  all cases, the pretrained model achieves better accuracy, 
and further, as a general trend, the larger the model the more of an impact pretraining has and the greater the improvement in classification performance.

With our experiments, we show that pretraining allows training larger (deeper) models that then perform almost equally good overall and can even outperform the initially favored smaller classifier networks in specific cases.

We provide pretrained weights for the DeepHS-Net+HyveConv as well as the ResNet-18+HyveConv model at \url{https://github.com/cogsys-tuebingen/hsi_benchmark}.

\section{Limitations and Future Work}
One obvious limitation of the proposed benchmark is the restriction to only three  hyperspectral applications. It is a significant improvement to the evaluation on the HRSS data set only, still the benchmark could be much more diverse. The intention is to enhance forthcoming iterations of this benchmark by encompassing a wider array of hyperspectral applications. Should you possess a suitable dataset for potential inclusion in the upcoming version, we encourage you to establish contact with the authors.

Besides that, the analysis of the results could be even more sophisticated. A more in depth investigation seems possible and also helpful for the design of better hyperspectral models. Future research will focus on this. 

\section{Conclusion}
We have established a comprehensive framework to assess models across various hyperspectral imaging (HSI) classification tasks. In pursuit of a universally applicable HSI classification model, we amalgamated three diverse datasets, each catering to distinct use cases, into a unified benchmark. This initiative aids in advancing the quest for a generalizing HSI classification model. Central to this benchmark is the establishment of fixed training and evaluation pipelines, thereby enabling impartial comparisons for future research.

In the context of the benchmark, 23 models were implemented and evaluated. The performance of these significantly different models is analyzed and discussed. As a result, we conclude, that the requirements on the model depend on the hyperspectral use case. Still, models considering the spatial and spectral dimension performed overall better. 
Moreover, a significant number of state-of-the-art models designed for HSI classification have been fine-tuned exclusively on the HRSS dataset. This dataset predominantly stresses the models' spatial feature extraction capabilities, with less emphasis on the spectral aspect. This observation underscores the necessity for a more comprehensive evaluation benchmark that encapsulates diverse scenarios. Additionally, our study highlights the substantial disparity in performance between patchwise and objectwise classification approaches employed by these models. Furthermore, we meticulously assessed the impact of the limited size of hyperspectral datasets, revealing distinct variations in how different models respond to this constraint.


To tackle the issue of the small data sets, we propose a pretraining strategy for hyperspectral image classifier models
and show that pretraining utilizing our hyperspectral benchmark can help to improve classification performance.
We claim that it is indeed helpful to combine the different HS data sets and train a shared backbone on multiple tasks, even if data sets and tasks differ considerably. It seems, we can even benefit from this large variety to extract the most general hyperspectral features.
A model backbone pretrained in this way stabilizes the subsequent training and reduces overfitting, which, in turn, enables to use larger (deeper) networks that are probably better suited to capture the complex spectral characteristics of hyperspectral image data.

We hope that this work makes future research easier and allows better comparison of different approaches. Further, we believe the insights reported here, and the described pretraining, will help to design better HSI models and allow the use of hyperspectral imaging in new areas.

\backmatter

\bmhead{Data availability and Code availability}
The benchmark data and code are released at \url{https://github.com/cogsys-tuebingen/hsi_benchmark}.
In addition, we provide pretrained model weights.

\bmhead{Acknowledgments}
Thanks to Manuel Graña for the permission to use the collection of the HRSS data set \citep{HRSSdatasets} for this benchmark.
This work has been supported by the German Ministry of Economy, Labour and Tourism, Project KI-PRO-BAU, FKZ: BW1\_2108/02.

\begin{appendices}

\section{Benchmark statistics}
\subsection{Training-validation-test set sizes}
\label{app:split}
Tab. \ref{tab:training_val_test_splits} enumerates all configurations within the proposed benchmark. For each configuration, the size of the training, validation and the test set is presented. The difference between objectwise (O) and patchwise (P) configurations is expectable, as a recording of a single object can be used to generate multiple patches.
\begin{table*}[]
    \centering
    \def\arraystretch{1.1}
    \begin{tabular}{llc|rrr}
    \hline
         Data set & Configuration & Type & Training set & Validation set & Test set \\
         \hline
            \multirow{11}{*}{\begin{tabular}{@{}l@{}} HRSS \end{tabular}} 
             &	Indian Pines with train ratio 0.05  & P &	385 &	120 &	9,744\\
             &	Indian Pines with train ratio 0.1  & P &	770 &	248 &	9,231\\
             &	Indian Pines with train ratio 0.3  & P &	2,307 &	760 &	7,182\\
             &	Pavia University with train ratio 0.05  & P &	1,605 &	530 &	40,641\\
             &	Pavia University with train ratio 0.1  & P &	3,208 &	1,065 &	38,503\\
             &	Pavia University with train ratio 0.3  & P &	9,625 &	3,204 &	29,947\\
             &	Salinas with train ratio 0.05 & P &	2,029 &	668 &	51,432\\
             &	Salinas with train ratio 0.1 & P &	4,059 &	1,344 &	48,726\\
             &	Salinas with train ratio 0.3 & P &	12,179 &	4,052 &	37,898 \\
            \hdashline
            \multirow{4}{*}{\begin{tabular}{@{}l@{}} DeepHS \\ Debris\end{tabular}} &	Objectwise with Coring HSI & O &	50 &	10 &	10\\
             &	Patchwise with Corning HSI& P &	7,624 &	1,570 &	1,292,409\\
             &	Objectwise with Specim FX 10 & O &	50 &	10 &	10\\
             &	Patchwise with Specim FX 10 & P &	5,974 &	1,262 &	1,119,347\\
            \hdashline
            \multirow{29}{*}{\begin{tabular}{@{}l@{}} DeepHS \\ Fruit \end{tabular}} &	
            \multirow{4}{*}{
                        Avocado \begin{tabular}{@{}lr@{}}
                        Firmness & Corning HSI \\
                        Firmness & Innospec RE \\
                        Firmness & Specim FX 10 \\
                        Ripeness & Corning HSI \\
                        Ripeness & Innospec RE \\
                        Ripeness & Specim FX 10 \\
                        \end{tabular}} & O &	50 &	9 &	9\\
             &	& O &	40 &	9 &	9\\
             &	& O &	139 &	23 &	24\\
             &	& O &	50 &	9 &	9\\
             &	& O &	40 &	9 &	9\\
             &	& O &	142 &	24 &	24\\
             &	 \multirow{4}{*}{
                        Kaki \begin{tabular}{@{}lr@{}}
                        Firmness & Corning HSI \\
                        Firmness & Specim FX 10 \\
                        Ripeness & Corning HSI \\
                        Ripeness & Corning HSI \\
                        Sugar & Specim FX 10 \\
                        Sugar & Specim FX 10 \\
                        \end{tabular}} & O &	56 &	12 &	12\\
             &	& O &	56 &	12 &	12\\
             &	& O &	56 &	12 &	12\\
             &	& O &	56 &	12 &	12\\
             &	& O &	56 &	12 &	12\\
             &	& O &	56 &	12 &	12\\
             &	\multirow{4}{*}{
                        Kiwi \begin{tabular}{@{}lr@{}}
                        Firmness & Innospec Redeye \\
                        Firmness & Specim FX 10 \\
                        Ripeness & Innospec Redeye \\
                        Ripeness & Specim FX 10 \\
                        Sugar & Innospec Redeye \\
                        Sugar & Specim FX 10 \\
                        \end{tabular}}& O &	58 &	9 &	9\\
             &	& O &	128 &	21 &	23\\
             &	& O &	58 &	9 &	9\\
             &	& O &	138 &	24 &	24\\
             &	& O &	58 &	9 &	9\\
             &	& O &	128 &	21 &	23\\
             &	\multirow{4}{*}{
                        Mango \begin{tabular}{@{}lr@{}}
                        Firmness & Corning HSI\\
                        Firmness & Specim FX 10 \\
                        Ripeness & Corning HSI \\
                        Ripeness & Specim FX 10\\
                        Sugar & Corning HSI \\
                        Sugar & Specim FX 10 \\
                        \end{tabular}} & O &	56 &	12 &	12\\
             &	& O &	56 &	12 &	12\\
             &	& O &	56 &	12 &	12\\
             &	& O &	56 &	12 &	12\\
             &	& O &	56 &	12 &	12\\
             &	& O &	56 &	12 &	12\\
             &	\multirow{4}{*}{
                        Papaya \begin{tabular}{@{}lr@{}}
                        Firmness & Corning HSI\\
                        Firmness & Specim FX 10 \\
                        Ripeness & Corning HSI \\
                        Ripeness & Specim FX 10\\
                        Sugar & Corning HSI \\
                        Sugar & Specim FX 10 \\
                        \end{tabular}}& O &	42 &	9 &	9\\
             &	& O &	42 &	9 &	9\\
             &	& O &	42 &	9 &	9\\
             &	& O &	42 &	9 &	9\\
             &	& O &	42 &	9 &	9\\
             &	& O &	42 &	9 &	9\\
    \hline
    \end{tabular}
    \caption{Sizes of the training, validation and test sets. Type 'P' indicates a patchwise task and 'O' an objectwise task. It is evident, that the patchwise cases have much more samples.}
    \label{tab:training_val_test_splits}
\end{table*}

\subsection{HRSS: Deterministic train-validation-test split}
\label{app:hrss_split}
As mentioned, the HRSS data set lacks an established definition of the training-validation-test sets. This hinders a fair comparison of results. We provide a solution with a variable ratio between the training set (containing pure training and validation set) and the test set. 

To establish deterministic sets for specific training-test ratios, we employ a pseudo-random pixel arrangement within each category. This approach guarantees the reproducibility of results, an essential prerequisite for enhancing research quality within this domain.

For a practical illustration of how to implement these splits, a comprehensive example is available at the following link: \url{https://github.com/cogsys-tuebingen/hsi_benchmark}.

\section{Hyperparameters}
Table \ref{tab:model_hparams} presents a comprehensive overview of the distinct hyperparameters associated with each model. Throughout the experimentation, the maximum feasible batch size, up to 32, was employed for all models. In the majority of cases, a uniform learning rate of $1 \times 10^{-3}$ was adopted, though deviations were permitted when the source literature indicated an alternative default learning rate. The identical principle governed the determination of training epochs and optimizer. 
\begin{table*}[]
    \centering
    \def\arraystretch{1.1}
    \begin{tabular}{l|rrrccc}
    \hline
         Model name & \begin{tabular}{@{}c@{}}Batch \\ size\end{tabular} & \# Epochs & \begin{tabular}{@{}c@{}}Learning \\ rate\end{tabular} & \begin{tabular}{@{}c@{}}Loss \\ function\end{tabular} & Optimizer & Scheduler\\
         \hline
         MLP & 32 & 50 & \textbf{0.001} & CE & Adam & Step-wise\\
         RNN & 32 & 50 & 0.01 & CE & Adam & Step-wise\\ 
         1D CNN & 32 & 50 & 0.01 & CE & Adam & Step-wise\\ 
         2D CNN & 32 & 50 & 0.01 & CE & Adam & Step-wise\\
         2D CNN (spatial) & 32 & 50 & 0.01 & CE & Adam & Step-wise\\
         2D CNN (spectral) & 32 & 50 & 0.01 & CE & Adam & Step-wise\\
         3D CNN & \textbf{8} & 50 & 0.01 & CE & Adam & Step-wise\\ 
         Gabor CNN & 32 & 50 & 0.01 & CE & Adam & Step-wise\\
         EMP CNN & 32 & 50 & 0.01 & CE & Adam & Step-wise\\ 
         ResNet-18 & 32 & 50 & 0.01 & CE & Adam & Step-wise\\
         ResNet-152 & 32 & 50 & 0.01 & CE & Adam & Step-wise\\
         ResNet-18+HyveConv & 32 & 50 & 0.01 & CE & Adam & Step-wise\\
         ResNet-152+HyveConv & 32 & 50 & 0.01 & CE & Adam & Step-wise\\ 
         DeepHS-Net & 32 & 50 & 0.01 & CE & Adam & Step-wise\\
         DeepHS-Net+HyveConv & 32 & 50 & 0.01 & CE & Adam & Step-wise\\ 
         DeepHS-Hybrid-Net & \textbf{4} & 50 & 0.01 & CE & Adam & Step-wise\\
         SpectralNET & \textbf{8} & 50 & 0.01 & CE & \textbf{SGD} & Step-wise\\
         HybridSN & \textbf{16} & 50 & \textbf{0.0001} & CE & Adam & Step-wise\\ 
         Attention-based CNN & 32 & 50 & \textbf{0.0001} & CE & Adam & Step-wise\\
         SpectralFormer & 32 & 50 & 0.01 & CE & Adam & Step-wise\\
         HiT & \textbf{16} & \textbf{100} & 0.01 & CE & Adam & Step-wise \\
    \hline
    \end{tabular}
    \caption{Model-specific hyperparameters used for training as described in Sec. \ref{sec:training_procedure}. Exceptions are marked in \textbf{bold}.}
    \label{tab:model_hparams}
\end{table*}






\end{appendices}


\bibliography{literature}

\begin{thebibliography}{41}
\providecommand{\natexlab}[1]{#1}
\providecommand{\url}[1]{{#1}}
\providecommand{\urlprefix}{URL }
\providecommand{\doi}[1]{\url{https://doi.org/#1}}
\providecommand{\eprint}[2][]{\url{#2}}
 \bibcommenthead

\bibitem[{Ahmad et~al(2022)Ahmad, Shabbir, Roy, Hong, Wu, Yao, Khan, Mazzara,
  Distefano, and Chanussot}]{AhmadSRHWYKMDC22}
Ahmad M, Shabbir S, Roy SK, et~al (2022) Hyperspectral image classification -
  traditional to deep models: {A} survey for future prospects. {IEEE} J Sel Top
  Appl Earth Obs Remote Sens 15:968--999. \doi{10.1109/JSTARS.2021.3133021},
  \urlprefix\url{https://doi.org/10.1109/JSTARS.2021.3133021}

\bibitem[{Barker and Rayens(2003)}]{barker2003partial}
Barker M, Rayens W (2003) Partial least squares for discrimination. Journal of
  Chemometrics: A Journal of the Chemometrics Society 17(3):166--173

\bibitem[{Bonifazi et~al(2019)Bonifazi, Capobianco, Serranti, and
  Palmieri}]{BonifaziCSP19}
Bonifazi G, Capobianco G, Serranti S, et~al (2019) Hyperspectral imaging
  applied to the waste recycling sector. Spectroscopy Europe 31:8--11.
  \doi{10.1255/sew.2019.a3}

\bibitem[{Chakraborty and Trehan(2021)}]{Chakraborty2021SpectralNETES}
Chakraborty T, Trehan U (2021) Spectralnet: Exploring spatial-spectral
  waveletcnn for hyperspectral image classification. ArXiv abs/2104.00341

\bibitem[{Chen et~al(2014)Chen, Lin, Zhao, Wang, and Gu}]{ChenLZWG14}
Chen Y, Lin Z, Zhao X, et~al (2014) Deep learning-based classification of
  hyperspectral data. IEEE Journal of Selected Topics in Applied Earth
  Observations and Remote Sensing 7(6):2094--2107.
  \doi{10.1109/JSTARS.2014.2329330}

\bibitem[{Cristianini and Shawe{-}Taylor(2010)}]{DBLP:books/daglib/0026018}
Cristianini N, Shawe{-}Taylor J (2010) An Introduction to Support Vector
  Machines and Other Kernel-based Learning Methods. Cambridge University Press

\bibitem[{Dosovitskiy et~al(2020)Dosovitskiy, Beyer, Kolesnikov, Weissenborn,
  Zhai, Unterthiner, Dehghani, Minderer, Heigold, Gelly, Uszkoreit, and
  Houlsby}]{DosovitskiyBKWZUDMHGUH20}
Dosovitskiy A, Beyer L, Kolesnikov A, et~al (2020) An image is worth 16x16
  words: Transformers for image recognition at scale. CoRR abs/2010.11929.
  \urlprefix\url{https://arxiv.org/abs/2010.11929},
  {\href{https://arxiv.org/abs/2010.11929}{{2010.11929}}}

\bibitem[{Fukushima(1980)}]{Fukushima80}
Fukushima K (1980) Neocognitron: A self-organizing neural network for a
  mechanism of pattern recognition unaffected by shift in position. Biological
  Cybernetics 36(4):193--202. \doi{10.1007/bf00344251},
  \urlprefix\url{https://doi.org/10.1007/bf00344251}

\bibitem[{George~Cybenko(1999)}]{Cybenko99}
George~Cybenko JRDianne P.~O'Leary (1999) The Mathematics of Information
  Coding, Extraction and Distribution. Springer New York

\bibitem[{Ghamisi et~al(2018)Ghamisi, Maggiori, Li, Souza, Tarablaka, Moser,
  De~Giorgi, Fang, Chen, Chi, Serpico, and Benediktsson}]{GhamisiMLSTM18}
Ghamisi P, Maggiori E, Li S, et~al (2018) New frontiers in spectral-spatial
  hyperspectral image classification: The latest advances based on mathematical
  morphology, markov random fields, segmentation, sparse representation, and
  deep learning. IEEE Geoscience and Remote Sensing Magazine 6(3):10--43.
  \doi{10.1109/MGRS.2018.2854840}

\bibitem[{Girod et~al(2008)Girod, Landry, Doyon, Osuna-Garcia, Salazar-Garcia,
  and Geonaga}]{GirodLDOSG08}
Girod D, Landry JA, Doyon G, et~al (2008) Evaluating hass avocado maturity
  using hyperspectral imaging. In: Caribbean Food Crops Society, 44th Annual
  Meeting, Caribbean Food Crops Society, \doi{10.22004/AG.ECON.256469},
  \urlprefix\url{https://ageconsearch.umn.edu/record/256469}

\bibitem[{Graña et~al(2011)Graña, Veganzons, and B.}]{HRSSdatasets}
Graña M, Veganzons MA, B. A (2011) Hyperspectral remote sensing scenes.
  \url{https://ehu.eus/ccwintco/index.php?title=Hyperspectral_Remote_Sensing_Scenes}

\bibitem[{He et~al(2020)He, Zhao, Yang, Zhang, and Li}]{HeZYZL20}
He J, Zhao L, Yang H, et~al (2020) {HSI-BERT:} hyperspectral image
  classification using the bidirectional encoder representation from
  transformers. {IEEE} Trans Geosci Remote Sens 58(1):165--178.
  \doi{10.1109/TGRS.2019.2934760},
  \urlprefix\url{https://doi.org/10.1109/TGRS.2019.2934760}

\bibitem[{He et~al(2016)He, Zhang, Ren, and Sun}]{HeZRS16}
He K, Zhang X, Ren S, et~al (2016) Deep residual learning for image
  recognition. In: 2016 {IEEE} Conference on Computer Vision and Pattern
  Recognition, {CVPR} 2016, Las Vegas, NV, USA, June 27-30, 2016. {IEEE}
  Computer Society, pp 770--778, \doi{10.1109/CVPR.2016.90},
  \urlprefix\url{https://doi.org/10.1109/CVPR.2016.90}

\bibitem[{Hong et~al(2022)Hong, Han, Yao, Gao, Zhang, Plaza, and
  Chanussot}]{HongHYGZPC22}
Hong D, Han Z, Yao J, et~al (2022) Spectralformer: Rethinking hyperspectral
  image classification with transformers. {IEEE} Trans Geosci Remote Sens
  60:1--15. \doi{10.1109/TGRS.2021.3130716},
  \urlprefix\url{https://doi.org/10.1109/TGRS.2021.3130716}

\bibitem[{Kendall(1957)}]{kendall1957course}
Kendall M (1957) A Course in Multivariate Analysis. Griffin's statistical
  monographs \& courses, Hafner Publishing Company,
  \urlprefix\url{https://books.google.de/books?id=8ROoAAAAIAAJ}

\bibitem[{Kingma and Ba(2015)}]{Kingma14}
Kingma DP, Ba J (2015) Adam: {A} method for stochastic optimization. In: Bengio
  Y, LeCun Y (eds) 3rd International Conference on Learning Representations,
  {ICLR} 2015, San Diego, CA, USA, May 7-9, 2015, Conference Track Proceedings,
  \urlprefix\url{http://arxiv.org/abs/1412.6980}

\bibitem[{Krizhevsky et~al(2012)Krizhevsky, Sutskever, and
  Hinton}]{KrizhevskySH12}
Krizhevsky A, Sutskever I, Hinton GE (2012) Imagenet classification with deep
  convolutional neural networks. In: Bartlett PL, Pereira FCN, Burges CJC,
  et~al (eds) Advances in Neural Information Processing Systems 25: 26th Annual
  Conference on Neural Information Processing Systems 2012. Proceedings of a
  meeting held December 3-6, 2012, Lake Tahoe, Nevada, United States, pp
  1106--1114,
  \urlprefix\url{https://proceedings.neurips.cc/paper/2012/hash/c399862d3b9d6b76c8436e924a68c45b-Abstract.html}

\bibitem[{LeCun(1987)}]{LeCun87}
LeCun Y (1987) Modéles connexionistes de l'apprentissage. PhD thesis

\bibitem[{Lee et~al(2018)Lee, Eum, and Kwon}]{Lee18}
Lee H, Eum S, Kwon H (2018) Cross-domain {CNN} for hyperspectral image
  classification. In: 2018 {IEEE} International Geoscience and Remote Sensing
  Symposium, {IGARSS} 2018, Valencia, Spain, July 22-27, 2018. {IEEE}, pp
  3627--3630, \doi{10.1109/IGARSS.2018.8519419},
  \urlprefix\url{https://doi.org/10.1109/IGARSS.2018.8519419}

\bibitem[{Lee et~al(2019)Lee, Eum, and Kwon}]{LeeEK19}
Lee H, Eum S, Kwon H (2019) Is pretraining necessary for hyperspectral image
  classification? In: 2019 {IEEE} International Geoscience and Remote Sensing
  Symposium, {IGARSS} 2019, Yokohama, Japan, July 28 - August 2, 2019. {IEEE},
  pp 3321--3324, \doi{10.1109/IGARSS.2019.8898734},
  \urlprefix\url{https://doi.org/10.1109/IGARSS.2019.8898734}

\bibitem[{Lee et~al(2022)Lee, Eum, and Kwon}]{LeeEK22}
Lee H, Eum S, Kwon H (2022) Exploring cross-domain pretrained model for
  hyperspectral image classification. {IEEE} Trans Geosci Remote Sens 60:1--12.
  \doi{10.1109/TGRS.2022.3165441},
  \urlprefix\url{https://doi.org/10.1109/TGRS.2022.3165441}

\bibitem[{Li et~al(2019)Li, Song, Fang, Chen, Ghamisi, and
  Benediktsson}]{LiSFCGB19}
Li S, Song W, Fang L, et~al (2019) Deep learning for hyperspectral image
  classification: An overview. {IEEE} Trans Geosci Remote Sens
  57(9):6690--6709. \doi{10.1109/TGRS.2019.2907932},
  \urlprefix\url{https://doi.org/10.1109/TGRS.2019.2907932}

\bibitem[{Lin et~al(2013)Lin, Chen, Zhao, and Wang}]{LinCZW13}
Lin Z, Chen Y, Zhao X, et~al (2013) Spectral-spatial classification of
  hyperspectral image using autoencoders. In: 9th International Conference on
  Information, Communications {\&} Signal Processing, {ICICS} 2013, Tainan,
  Taiwan, December 10-13, 2013. {IEEE}, pp 1--5,
  \doi{10.1109/ICICS.2013.6782778},
  \urlprefix\url{https://doi.org/10.1109/ICICS.2013.6782778}

\bibitem[{Lorenzo et~al(2020)Lorenzo, Tulczyjew, Marcinkiewicz, and
  Nalepa}]{LorenzoTMN20}
Lorenzo PR, Tulczyjew L, Marcinkiewicz M, et~al (2020) Hyperspectral band
  selection using attention-based convolutional neural networks. {IEEE} Access
  8:42384--42403. \doi{10.1109/ACCESS.2020.2977454},
  \urlprefix\url{https://doi.org/10.1109/ACCESS.2020.2977454}

\bibitem[{Lu et~al(2020)Lu, Dao, Liu, He, and Shang}]{LuDLHS20}
Lu B, Dao P, Liu J, et~al (2020) Recent advances of hyperspectral imaging
  technology and applications in agriculture. Remote Sensing 12(16):2659.
  \doi{10.3390/rs12162659}, \urlprefix\url{https://doi.org/10.3390/rs12162659}

\bibitem[{Lu and Fei(2014)}]{LuF14}
Lu G, Fei B (2014) Medical hyperspectral imaging: {A} review. Journal of
  Biomedical Optics 19(1):010901. \doi{10.1117/1.jbo.19.1.010901},
  \urlprefix\url{https://doi.org/10.1117/1.jbo.19.1.010901}

\bibitem[{Mou et~al(2017)Mou, Ghamisi, and Zhu}]{MouGZ17}
Mou L, Ghamisi P, Zhu XX (2017) Deep recurrent neural networks for
  hyperspectral image classification. {IEEE} Trans Geosci Remote Sens
  55(7):3639--3655. \doi{10.1109/TGRS.2016.2636241},
  \urlprefix\url{https://doi.org/10.1109/TGRS.2016.2636241}

\bibitem[{Paoletti et~al(2019)Paoletti, Haut, Plaza, and Plaza}]{PaolettiHPP19}
Paoletti ME, Haut JM, Plaza J, et~al (2019) Deep learning classifiers for
  hyperspectral imaging: A review. Isprs Journal of Photogrammetry and Remote
  Sensing 158:279--317

\bibitem[{Pearson(1901)}]{doi:10.1080/14786440109462720}
Pearson K (1901) Liii. on lines and planes of closest fit to systems of points
  in space. The London, Edinburgh, and Dublin Philosophical Magazine and
  Journal of Science 2(11):559--572. \doi{10.1080/14786440109462720}

\bibitem[{Prechelt(1998)}]{Prechelt98}
Prechelt L (1998) Automatic early stopping using cross validation: quantifying
  the criteria. Neural Networks 11(4):761--767.
  \doi{https://doi.org/10.1016/S0893-6080(98)00010-0},
  \urlprefix\url{https://www.sciencedirect.com/science/article/pii/S0893608098000100}

\bibitem[{Roy et~al(2020)Roy, Krishna, Dubey, and Chaudhuri}]{RoyKDC20}
Roy SK, Krishna G, Dubey SR, et~al (2020) Hybridsn: Exploring 3-d-2-d {CNN}
  feature hierarchy for hyperspectral image classification. {IEEE} Geosci
  Remote Sens Lett 17(2):277--281. \doi{10.1109/LGRS.2019.2918719},
  \urlprefix\url{https://doi.org/10.1109/LGRS.2019.2918719}

\bibitem[{Rumelhart et~al(1986)Rumelhart, Hinton, and Williams}]{RumelhartHW86}
Rumelhart DE, Hinton GE, Williams RJ (1986) Learning representations by
  back-propagating errors. Nature 323(6088):533--536. \doi{10.1038/323533a0},
  \urlprefix\url{https://doi.org/10.1038/323533a0}

\bibitem[{Russakovsky et~al(2014)Russakovsky, Deng, Su, Krause, Satheesh, Ma,
  Huang, Karpathy, Khosla, Bernstein, Berg, and Fei{-}Fei}]{Russakovsky14}
Russakovsky O, Deng J, Su H, et~al (2014) {ImageNet} large scale visual
  recognition challenge. CoRR abs/1409.0575.
  \urlprefix\url{http://arxiv.org/abs/1409.0575},
  {\href{https://arxiv.org/abs/1409.0575}{{1409.0575}}}

\bibitem[{Varga et~al(2021)Varga, Makowski, and Zell}]{VargaMZ21}
Varga LA, Makowski J, Zell A (2021) Measuring the ripeness of fruit with
  hyperspectral imaging and deep learning. In: International Joint Conference
  on Neural Networks, {IJCNN} 2021, Shenzhen, China, July 18-22, 2021. {IEEE},
  pp 1--8, \doi{10.1109/IJCNN52387.2021.9533728},
  \urlprefix\url{https://doi.org/10.1109/IJCNN52387.2021.9533728}

\bibitem[{Varga et~al(2023{\natexlab{a}})Varga, Frank, and Zell}]{VargaFZ23}
Varga LA, Frank H, Zell A (2023{\natexlab{a}}) Self-supervised pretraining for
  hyperspectral classification of fruit ripeness. In: 6th International
  Conference on Optical Characterization of Materials, {OCM} 2023, Karlsruhe,
  Germany, March 22 - 23, 2023. {KIT Scientific Publishing}, pp 97--108

\bibitem[{Varga et~al(2023{\natexlab{b}})Varga, Messmer, Benbarka, and
  Zell}]{VargaMBZ23}
Varga LA, Messmer M, Benbarka N, et~al (2023{\natexlab{b}}) Wavelength-aware 2d
  convolutions for hyperspectral imaging. In: {IEEE/CVF} Winter Conference on
  Applications of Computer Vision, {WACV} 2023, Waikoloa, HI, USA, January 2-7,
  2023. {IEEE}, pp 3777--3786, \doi{10.1109/WACV56688.2023.00378},
  \urlprefix\url{https://doi.org/10.1109/WACV56688.2023.00378}

\bibitem[{Windrim et~al(2018)Windrim, Melkumyan, Murphy, Chlingaryan, and
  Ramakrishnan}]{WindrimMMCR18}
Windrim L, Melkumyan A, Murphy RJ, et~al (2018) Pretraining for hyperspectral
  convolutional neural network classification. {IEEE} Trans Geosci Remote Sens
  56(5):2798--2810. \doi{10.1109/TGRS.2017.2783886},
  \urlprefix\url{https://doi.org/10.1109/TGRS.2017.2783886}

\bibitem[{Yan et~al(2010)Yan, Zhao, Xue, Kou, and Liu}]{Yan10}
Yan Y, Zhao Y, Xue Hf, et~al (2010) Integration of spatial-spectral information
  for hyperspectral image classification. In: 2010 Second IITA International
  Conference on Geoscience and Remote Sensing, IEEE, pp 242--245

\bibitem[{Yang et~al(2018)Yang, Ye, Li, Lau, Zhang, and Huang}]{YangYLLZH18}
Yang X, Ye Y, Li X, et~al (2018) Hyperspectral image classification with deep
  learning models. {IEEE} Trans Geosci Remote Sens 56(9):5408--5423.
  \doi{10.1109/TGRS.2018.2815613},
  \urlprefix\url{https://doi.org/10.1109/TGRS.2018.2815613}

\bibitem[{Yang et~al(2022)Yang, Cao, Lu, and Zhou}]{YangCLZ22}
Yang X, Cao W, Lu Y, et~al (2022) Hyperspectral image transformer
  classification networks. {IEEE} Trans Geosci Remote Sens 60:1--15.
  \doi{10.1109/TGRS.2022.3171551},
  \urlprefix\url{https://doi.org/10.1109/TGRS.2022.3171551}

\end{thebibliography}

\end{document}